\documentclass[sigconf]{acmart}
\AtBeginDocument{%
  }

\copyrightyear{2026}
\acmYear{2026}
\setcopyright{cc}
\setcctype{by}
\acmConference[MM '26]{Proceedings of the 34th ACM International Conference on Multimedia}{November 10--14, 2026}{Rio de Janeiro, Brazil}
\acmBooktitle{Proceedings of the 34th ACM International Conference on Multimedia (MM '26), November 10--14, 2026, Rio de Janeiro, Brazil}
\acmDOI{10.1145/3767308.3836009}
\acmISBN{979-8-4007-2213-4/2026/11}

\acmSubmissionID{6036}

\usepackage{xspace}
\usepackage{microtype}
\usepackage{multirow}
\usepackage{fontawesome5}
\usepackage{makecell}
\usepackage{algorithm}
\usepackage{algorithmic}
\usepackage{tcolorbox}
\usepackage{hyperref}
\usepackage{cleveref}

\newcommand{\frameworkname}{\textsc{UrbanWorld2.0}\xspace}
\begin{document}

\title{\frameworkname: A Multimodal Agentic Framework for Reality-Aligned 3D World Generation at City-Scale}

\author{Shengyuan Wang}
\authornote{These authors contributed equally to this work.}
\orcid{0009-0000-3908-6153}
\affiliation{%
  \department{College of AI}
  \institution{Tsinghua University}
  \city{Beijing}
  \country{China}
}
\email{wangshengyuan25@mails.tsinghua.edu.cn}

\author{Zhiheng Zheng}
\authornotemark[1]
\orcid{0009-0007-5740-6801}
\affiliation{%
  \department{Shenzhen International Graduate School}
  \institution{Tsinghua University}
  \city{Shenzhen}
  \state{Guangdong}
  \country{China}
}
\email{zhengzhi24@mails.tsinghua.edu.cn}

\author{Yu Shang}
\orcid{0009-0005-9049-8483}
\affiliation{%
  \department{Department of Electronic Engineering}
  \institution{Tsinghua University}
  \city{Beijing}
  \country{China}
}
\email{shangy2021@gmail.com}

\author{Lixuan He}
\orcid{0009-0007-4596-5577}
\affiliation{%
  \department{Department of Electronic Engineering}
  \institution{Tsinghua University}
  \city{Beijing}
  \country{China}
}
\email{helx23@mails.tsinghua.edu.cn}

\author{Yangcheng Yu}
\orcid{0009-0008-3709-9872}
\affiliation{%
  \department{Department of Electronic Engineering}
  \institution{Tsinghua University}
  \city{Beijing}
  \country{China}
}
\email{yuyc23@mails.tsinghua.edu.cn}

\author{Hangyu Fan}
\orcid{0009-0007-1409-8706}
\affiliation{%
  \department{Department of Electronic Engineering}
  \institution{Tsinghua University}
  \city{Beijing}
  \country{China}
}
\email{fanhangyu@mail.tsinghua.edu.cn}

\author{Jie Feng}
\authornote{Corresponding authors.}
\orcid{0000-0003-3279-7117}
\affiliation{%
  \institution{Zhongguancun Academy}
  \city{Beijing}
  \country{China}
}
\email{fengjie@bza.edu.cn}

\author{Qingmin Liao}
\orcid{0000-0002-7509-3964}
\affiliation{%
  \department{Shenzhen International Graduate School}
  \institution{Tsinghua University}
  \city{Shenzhen}
  \state{Guangdong}
  \country{China}
}
\email{liaoqm@sz.tsinghua.edu.cn}

\author{Yong Li}
\authornotemark[2]
\orcid{0000-0001-5617-1659}
\affiliation{%
  \department{Department of Electronic Engineering, BNRist}
  \institution{Tsinghua University}
  \city{Beijing}
  \country{China}
}
\email{liyong07@tsinghua.edu.cn}

\renewcommand{\shortauthors}{Wang et al.}

\begin{abstract}

The automated generation of high-fidelity, city-scale 3D environments remains a formidable challenge with profound academic and industrial implications. However, existing methods struggle to achieve the necessary quality, fidelity, and scalability. To address this, we propose \frameworkname, a reality-aligned intelligent multimodal synthesis engine that creates detailed, city-scale 3D worlds of high fidelity. We introduce an agentic framework that leverages diverse multimodal foundation tools to acquire real-world knowledge, maintain robust intermediate representations, and construct complex 3D scenes. 
This agentic design, featuring dynamic data processing, iterative self-reflection and refinement, and the invocation of advanced multimodal tools, minimizes cumulative errors and enhances overall performance. Extensive quantitative experiments and qualitative analyzes validate the superior performance of \frameworkname in real-world alignment, shape precision, texture fidelity, and aesthetics level, achieving a win rate of over 86\% against existing baselines for overall perceptual quality. This combination of 3D quality, reality alignment, scalability, and seamless compatibility with computer graphics pipelines makes \frameworkname a promising foundation for applications in immersive media, embodied intelligence, and world models.

\end{abstract}

\begin{teaserfigure}
  \includegraphics[width=\textwidth]{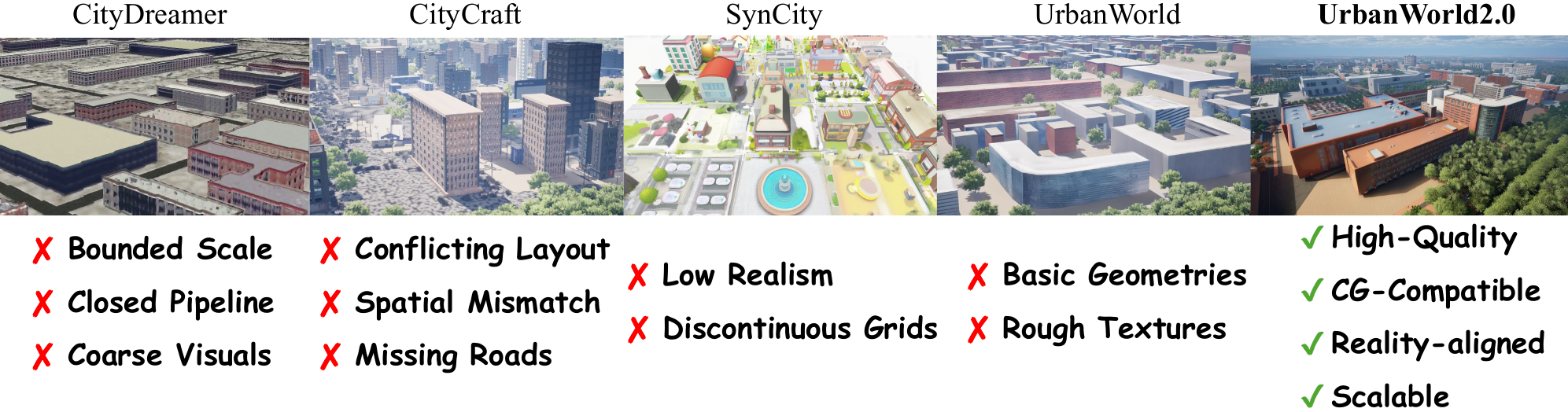}
  \caption{\frameworkname is a solution to challenges regarding quality, fidelity, scalability and compatibility in 3D urban world generation.}
  \Description{}
  \label{fig:teaser}
\end{teaserfigure}

\maketitle

\section{Introduction}
\label{sec:intro}

\begin{figure*}[htbp]
    \centering
    \includegraphics[width=\linewidth]{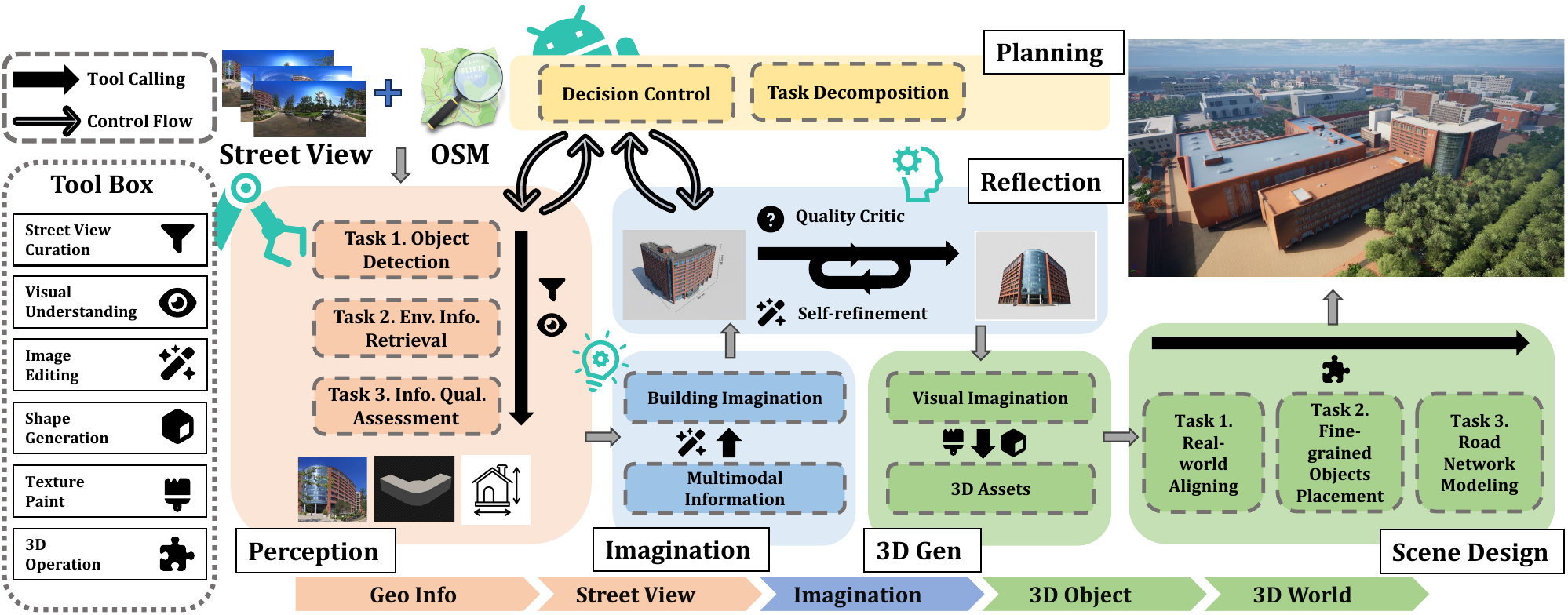}    
    \caption{\frameworkname is a multimodal agentic framework for generating high-quality, reality-aligned 3D urban worlds at city-scale. }
    \label{fig:framework}
\end{figure*}

The generation of high-quality 3D worlds represents a critical research frontier with profound implications for immersive media~\cite{anantrasirichai2022artificial,lee2024all,lavalle2023virtual,soliman2024artificial}, large-scale simulation~\cite{zhou2025virtualcommunityopenworld, piao2025agentsociety}, and the development of embodied intelligence~\cite{wang2024grutopia,wu2024metaurban,parish2001procedural,black2024pi_0} or world models~\cite{ding2024understanding,liu2025generative,agarwal2025cosmos,zhu2024sora}. However, the creation of such content, particularly the creation of complex 3D scenes, remains a predominantly labor-intensive process~\cite{wen20253dscenegenerationsurvey}. This reliance on manual creation is both cost-prohibitive and time-consuming, presenting a significant bottleneck to achieving greater scale and quality, and necessitating the development of automated solutions.

Indoor scenes represent a main category of 3D environments~\cite{raistrick2024infinigen,tang2024diffuscene,hollein2023text2room}, as they are relevant to daily life and relatively easy to capture from the real world. However, existing 3D indoor scene scenarios are often limited in scale and complexity, which creates a significant domain gap for many applications requiring large-scale or highly heterogeneous data. The generation of 3D urban environments~\cite{parish2001procedural,lin2023infinicity} emerges as a key research thrust. As highly complex systems, cities are characterized by intricate spatial layouts and a high degree of component heterogeneity, such as diverse architectural forms. Consequently, the ability to model or reconstruct entire 3D cities is indispensable for applications with substantial industrial potential, including urban simulation, autonomous driving, and embodied AI. A primary objective in this context is the creation of near-realistic or reality-aligned urban worlds. Such fidelity is crucial for minimizing the sim-to-real gap and mitigating the costs associated with domain adaptation.

While existing works~\cite{shang2024urbanworld,engstler2025syncity,deng2024citycraft,xie2024citydreamer,shen2022sgam} have endeavored to advance 3D world construction, they face several significant and unresolved challenges. First, the large scale of urban 3D scenes, often comprising thousands of individual objects, \textbf{imposes prohibitive computational costs} for both training and generation. This is especially problematic for methods reliant on visual- or neural 3D- based representations. Second, many approaches are \textbf{constrained in generation quality by data and input limitations}. They often utilize simplistic environmental information or omit it entirely, while instructions guiding 3D generation are typically restricted to text, thus neglecting the richer context available from multimodal inputs. Compounding this issue is the \textbf{limited availability and quality of real-world data}, such as imperfect imagery or insufficient and imbalanced GIS datasets. Finally, accurately representing the complex layout of a real city remains a major hurdle. The simultaneous generation of diverse urban elements (e.g.,  buildings, roads, and green spaces) is inherently difficult. Moreover, many current methods \textbf{lack capacity for real-world layout retrieval and object alignment}, which are vital for creating reality-aligned 3D worlds. 

Recently, the increasing power of multimodal generative foundation models has enabled the development of sophisticated agentic frameworks~\cite{garg2025designing} for 3D world generation. 
Agentic core components, such as planning, execution with tools, and iterative self-reflection, provide a novel mechanism for addressing the complexity and scale of this task. This offers a path to overcome the scalability and fidelity constraints of the field.

In response to these fundamental limitations, we propose \frameworkname, a multimodal agentic framework designed to automatically create high-fidelity, scalable, and reality-aligned 3D worlds from real-world environmental information. First, we introduce an agent-based, training-free methodology that fully leverages multimodal tools, thereby eliminating prohibitive training costs. The automatic generation process is intentionally modular, a design choice that significantly enhances controllability and parallelism. The mesh-based 3D representation and \frameworkname architecture are not only compatible with established computer graphics (CG) pipelines but are also demonstrably \textbf{scalable}. Second, to enhance the fidelity and quality of the generated 3D world, \frameworkname implements a sophisticated process for the automatic selection and curation of real-world data. This multi-stage process, in conjunction with the tool utilization by multimodal foundation models, elevates the quality and usability of the raw data, thereby establishing a robust foundation for \textbf{high-quality} 3D world generation. Third, the element-by-element procedure, coupled with layout control guided by real-world geospatial information, serves to mitigate the risk of misalignment. Furthermore, the integrated reflection-refinement mechanism contributes to the improved quality and accuracy of the generated 3D objects. \frameworkname consequently achieves \textbf{superior reality-alignment }in comparison to contemporary methods. This overall approach also integrates fine-grained modeling of environmental details, such as urban elements and traffic patterns, which further enhances the realism, fidelity, and utility of the generated 3D worlds.
In summary, our main contributions are as follows:
\begin{itemize}
    \item We propose \frameworkname, a novel multimodal agentic framework for generating city-scale, reality-aligned 3D worlds. Its agentic design overcomes fundamental challenges in quality, fidelity, robustness, and scalability, providing a solid and reality-aligned foundation for important downstream applications.
    \item Through comprehensive evaluation, we show that \frameworkname achieves SOTA performance in urban 3D world generation, excelling in key metrics including shape precision, texture fidelity, reality alignment, consistency, and compatibility with standard CG pipelines.
    \item The source codes and all generated 3D urban world assets are open-sourced for the community, enabling continuous optimization and broader application.

\end{itemize}

\definecolor{iconblue}{HTML}{2980B9}
\definecolor{icongreen}{HTML}{27AE60}
\definecolor{iconorange}{HTML}{F39C12}
\definecolor{iconpurple}{HTML}{8E44AD}
\definecolor{iconred}{HTML}{C0392B}
\definecolor{icongray}{HTML}{7F8C8D}
\definecolor{iconcyan}{HTML}{17a2b8}

\newcommand{\roleprep}{\textcolor{icongray}{\faFilter\ }}
\newcommand{\rolepercept}{\textcolor{iconblue}{\faEye\ }}
\newcommand{\roleimagine}{\textcolor{iconpurple}{\faMagic\ }}
\newcommand{\rolereflect}{\textcolor{iconorange}{\faQuestionCircle\ }}
\newcommand{\rolegen}{\textcolor{iconred}{\faCube\ }}
\newcommand{\rolepost}{\textcolor{icongreen}{\faPuzzlePiece\ }}
\newcommand{\roledesign}{\textcolor{iconcyan}{\faDraftingCompass\ }}

\definecolor{modalT}{RGB}{214, 234, 248}    %
\definecolor{modalI}{RGB}{252, 228, 214}    %
\definecolor{modal3D}{RGB}{212, 239, 223}   %
\definecolor{modalTgeo}{RGB}{253, 235, 208} %

\newlength{\fixedmodalwidth}
\newlength{\fixedmodalheight}
\newlength{\fixedmodaldepth}

\setlength{\fixedmodalwidth}{1.6ex}  %
\setlength{\fixedmodalheight}{1.6ex} %
\setlength{\fixedmodaldepth}{0ex}  %

\newcommand{\modbox}[2]{{%
  \scalebox{0.75}{%
    \fcolorbox{#1}{#1}{%
      \makebox[\fixedmodalwidth][c]{\raisebox{0pt}[\fixedmodalheight][\fixedmodaldepth]{\sffamily #2}}%
    }%
  }%
}}

\newcommand{\modT}{\modbox{modalT}{T}}
\newcommand{\modI}{\modbox{modalI}{I}}

\newcommand{\modThreeD}{\modbox{modal3D}{3D}}

\newcommand{\modTgeo}{\modbox{modalTgeo}{T\textsubscript{G}}}

\begin{table*}[htbp]
\centering

  \caption{
  The \frameworkname toolbox, detailing each component's \textbf{Role}, \textbf{Task}, \textbf{Backbone}, and \textbf{Modality} (\modT{}~Text, \modI{}~Image, \modThreeD{}~3D, \modTgeo{}~Geospatial Information in Text).
}

\label{tab:tool_box}

\begin{tabular}{llcl} 
\toprule
\textbf{Role} & \textbf{Task} & \textbf{Modality} & \textbf{Backbone} \\
\midrule

\multirow{2}{*}{\roleprep Preparation} 
 & Geospatial Info. Retrieval & \modTgeo{} \modThreeD{} & \texttt{OSM API} \\
 & Street View Curation & \modTgeo{} \modI{} & \texttt{OSM API, Google/Baidu Maps API} \\
\addlinespace[2pt]

\multirow{3}{*}{\rolepercept Perception} 
 & Object Detection & \modI{} & \texttt{owlvit-base-patch32} \\
 & Env. Info. Retrieval & \modT{} \modI{} & \texttt{Qwen2.5-VL-72B-Instruct} \\
 & Quality Evaluation & \modT{} \modI{} & \texttt{gpt-5-mini} \\
\addlinespace[2pt]

\roleimagine Imagination 
 & Image Generation, Image Editing & \modT{} \modI{} & \texttt{gemini-2.5-flash-image} \\
\addlinespace[2pt]

\rolereflect Reflection 
 & Visual Question Answering & \modT{} \modI{} & \texttt{gpt-5-mini} \\
\addlinespace[2pt]

\multirow{2}{*}{\rolegen 3D Generation} 
 & Shape Generation & \modThreeD{} \modI{} & \texttt{Hunyuan3D-DiT} \\
 & Texture Paint & \modThreeD{} \modI{} & \texttt{Hunyuan3D-Paint} \\
\addlinespace[2pt]

 \multirow{3}{*}{\roledesign Scene Design} 
 & Real-world Alignment & \modTgeo{} \modThreeD{} & \texttt{Blender, OSM API} \\
 & Fine-grained Object Placement & \modThreeD{} & \texttt{Blender} \\
 & Road Network Modeling & \modThreeD{} & \texttt{Blender, MOSS} \\

\bottomrule
\end{tabular}
\end{table*}

\section{Methods} \label{sec:methods}
\subsection{Overview}
As illustrated in Figure \ref{fig:framework}, \frameworkname orchestrates the entire pipeline across six stages. The first stage is \emph{Planning}, which includes task decomposition and decision control. Subsequently, the \emph{Perception} stage queries, reviews, selects, and processes visual and geographic source data, establishing a robust foundation for the reality and fidelity of generation. Next, the \emph{Imagination} stage refines the intermediate representations, addressing and rectifying imperfections from the initial perception results. This is followed by the \emph{Reflection} stage, where the agent conducts iterative self-critique and self-refinement for quality control. The \emph{3D Gen} stage includes the actions of calling 3D generative tools to create high-quality 3D assets. Finally, the \emph{Scene Design} stage assembles the complete 3D world. This involves executing real-world alignment for all 3D elements and simulating fine-grained urban details, including object placement, road network generation, and traffic modeling. A suite of multimodal tools is utilized throughout this process, as detailed in Table \ref{tab:tool_box}.

\subsection{Task Planning}
The \emph{Planning} phase is structured around two principal components: task decomposition and decision control. To address the overarching objective of generating a 3D urban world from geographical coordinates, \frameworkname initially subdivides this complex undertaking into five distinct stages: \emph{Perception}, \emph{Imagination}, \emph{Reflection}, \emph{3D Gen}, and \emph{Scene Design}. Such a modular operational architecture significantly enhances both flexibility and controllability by effectively disentangling the heterogeneous sub-tasks inherent in the generation of complex 3D worlds. During the execution of the \emph{Perception} and \emph{Reflection} stages, the planning module generates control signals in response to feedback received from integrated tools. For instance, within the \emph{Perception} stage, object detection models are utilized for the analysis of the input image, a process that yields object candidates and their corresponding confidence scores. These scores subsequently serve as a quantitative basis for decision-making concerning the input images. Analogously, in the \emph{Reflection} stage, the determination of whether an imagined image is qualified for subsequent procedures is contingent upon an assessment by a quality evaluator. The intricate mechanisms governing each of these stages are elaborated upon in the following sections.

\subsection{Elements Perception}
Creating 3D worlds aligned with reality depends on reliable and scalable real-world data. Our approach leverages two key sources: structured geospatial data from OpenStreetMap (OSM)~\cite{OpenStreetMap} and panoramic street view imagery. The process begins by obtaining geographic details of buildings and other urban elements from OSM for a target location. The agent then acquires corresponding street view panoramas using an Online Map API. However, these images are often cluttered with extraneous elements such as irrelevant vegetation, construction, and vehicles, which can hinder the generation process. To address this, the agent first removes the obstacles and segments the raw images for buildings using an object detection model~\cite{minderer2022simple}. It then decides on most suitable image for subsequent steps based on model's judgment.

\subsection{Imagination and Reflection}

\subsubsection{Imaging Target Building}
Street-view images are a valuable source of information for describing building facades and the surrounding urban environment. However, this data source presents non-negligible limitations. Primarily, the constrained viewpoints fail to capture the entirety of a building's features, particularly its three-dimensional spatial and volumetric characteristics. Additionally, the presence of transient obstacles (e.g. vehicles, vegetation, and construction sites) during image acquisition frequently results in occlusions, which obscure parts of the building and pose a significant challenge to the construction of accurate 3D models.

To overcome these challenges, we propose a novel approach inspired by the human cognitive ability to form a complete mental image from incomplete sensory data~\cite{shepard1978mental,kunda2018visual}. This method introduces a computational process of ``imagining" the target building, which serves to create a more holistic representation of the structure and inform the 3D generation function. Visualized representation contains rich spatial, structural, and textural information, which becomes an informative and effect representation of mental image. The successful reconstruction of an entire building from such limited visual instruction necessitates a foundation of extensive world knowledge and sophisticated reasoning capabilities. \frameworkname provides the agent with a tool interface to gemini-2.5-flash-image~\cite{Google2025GeminiOverview}, a large pretrained multimodal foundation model, which the agent uses to analyze, imagine, repair, and reconstruct the building in the visual space.

Introducing other information sources like its overall geometry structure and geographical volume is another promising approach to enhance the accurate imagination of a building with extra clues about the target. Following successful practice of a prior work~\cite{shang2024urbanworld}, we develop a geographical information retrieval pipeline based on OSM, the high-quality and updating open-source geography service. The spatial spanning and rough geometry of a building are extracted and processed. Then, these different kinds of multi-modal information are taken as inputs by the agent along with street-view images.

\subsubsection{Reflecting and Refining}
To address the challenge of cumulative error propagation and improve the fidelity of the final output, we augment our generative agent with a reflection and refinement mechanism. This mechanism is implemented as a module driven by a vision-language model (VLM), which serves as an automated quality critic. Upon generating an initial building representation, the VLM-powered agent performs a preliminary quality assessment using a set of overall quality evaluation guidelines. Images scoring below a predetermined threshold are designated as candidates for an iterative refinement process. During this process, each candidate is subjected to a deeper evaluation, focusing on its \textbf{semantic plausibility}, \textbf{structural integrity}, and \textbf{aesthetic appearance}. Weakness reports and improvement guidance are also generated by the agent to instruct the next-step regeneration. The image is then regenerated and re-assessed in a loop, which terminates when generated image achieves the required quality score or the maximum attempts is exhausted.

\subsection{3D Building Assets Generation}
The generated and refined 2D building images from the preceding stage contain rich structural and appearance information, serving as high-quality visual prompts for 3D generation. 
To execute this stage, the agent is equipped with a specialized toolset derived from the Hunyuan-3D suite~\cite{hunyuan3d2025hunyuan3d} for its high-fidelity, visual-conditioned generation and open-source availability. First, the agent employs Hunyuan-3D-Dit-v2.1~\cite{hunyuan3d2025hunyuan3d} to generate an untextured 3D mesh from corresponding visual representations. Upon receiving the base shape, Hunyuan3D-Paint-v2.1~\cite{hunyuan3d2025hunyuan3d} is leveraged to synthesize a high-quality texture map. 
While these 3D generation tools are powerful, their raw outputs may contain topological errors that could interfere with subsequent procedures. A post-processing pipeline is thus executed to refine the 3D object. This process involves detecting and removing extraneous artifacts, such as unwanted ground planes and geometrically outlying fragments. Having successfully generated and refined the asset, the agent now holds a clean, high-fidelity, and textured 3D asset. It then concludes this stage by passing this object to the final \emph{Scene Design} stage for world integration.

\subsection{3D Scene Design}

\subsubsection{Real-world Aligning}
Beyond the fidelity of individual 3D assets, a significant challenge in large-scale world generation lies in the coherent spatial arrangement of these objects. For creating reality-aligned 3D scenes, the precise placement of each component is a critical determinant of overall quality and suitability for downstream applications. To address this, we introduce a framework that utilizes map data to systematically position high-quality, pre-generated building models.

Our method begins by extracting geospatial metadata from OSM, including the locations and attributes of diverse urban elements such as roads, building footprints, vegetation, and water bodies. This information is used to render a low-fidelity, schematic 3D scene that serves as a foundational scaffold for the final composition. Subsequently, each high-quality 3D object is integrated into this scaffold by aligning its key attributes: its \textbf{position} is directly inherited from the corresponding entity in the reference scene; its \textbf{scale} is uniformly adjusted according to the volume ratio between the target object and its reference counterpart; and its \textbf{orientation} is set by rotating the model to the angle that maximizes the ground-plane footprint overlap with the reference building footprint. Apart from the buildings, other urban elements including roads, vegetation, water bodies are introduced into the world and placed according to their location information from metadata. Ground and sky are rendered with existing assets.

\begin{table*}[htbp]
\centering
\caption{Performance comparison of our method against representative city-scale 3D generation approaches, demonstrating its competitiveness for both \textbf{region-level layout accuracy} and \textbf{street-level visual quality}. *``vs. Ours" indicates the percentage of time a baseline was preferred over \frameworkname, while ``vs. Baselines" reports the average win rate of each method against all methods. $\uparrow$ (higher is better) and $\downarrow$ (lower is better) indicate preferred metric directions. \textbf{Bold} and \underline{underlined} values denote the best and second-best methods.}
\label{tab:quantitative_exp}
\resizebox{0.95\textwidth}{!}{
\begin{tabular}{l cc c c c c c} 
\toprule
\textbf{Method} & \multicolumn{3}{c}{\textbf{Region-Level}} & \multicolumn{4}{c}{\textbf{Street-Level}}
\\
\cmidrule(lr){2-4} \cmidrule(lr){5-8} %
& \multicolumn{2}{c}{\small Layout} & \small Consistency & \small Aesthetics & \small vs. Ours & \small vs. Baselines & \\
\cmidrule(lr){2-3} \cmidrule(lr){4-4} \cmidrule(lr){5-5} \cmidrule(lr){6-6} \cmidrule(lr){7-7} %
& LPIPS $\downarrow$ & E-IoU $\uparrow$ & \makecell{Subject \\ Consist. $\uparrow$} & \makecell{LAP Score $\uparrow$} & Win Rate $\uparrow$ & Win Rate $\uparrow$ & \makecell{GPT-5.4 \\ Score $\uparrow$} \\ %
\midrule
SGAM~\cite{shen2022sgam} & 0.7179 & 0.0314 & - & - & - & - & - \\ 
CityDreamer~\cite{xie2025citydreamer4d}  & 0.6053 & 0.0675 & 0.9512 & 4.7412 & 2.0\% & 45.0\% & 3.70 \\ 
CityDreamer4D~\cite{xie2024citydreamer}  & 0.6006 & \textbf{0.0795} & \underline{0.9557} & 5.3716 & 0.0\% & 39.3\% & 3.45 \\ 
CityCraft~\cite{deng2024citycraft} & 0.6665 & 0.0573 & 0.9496 & 5.6338 & 3.0\% & 50.4\% & 4.35 \\ 
UrbanWorld~\cite{shang2024urbanworld} & \textbf{0.5231} & 0.0681 & 0.9469 & 4.7303 & \underline{14.0\%} & \underline{55.2\%} & \underline{5.25} \\ 
SynCity~\cite{engstler2025syncity} & 0.6862 & 0.0572 & \textbf{0.9781} & \underline{5.8204} & 3.0\% & 6.8\% & 2.80 \\
\midrule
\textbf{Ours} & \underline{0.5487} & \underline{0.0784} & 0.9524 & \textbf{5.9833} & \textbf{-}* & \textbf{86.9\%} & \textbf{6.05} \\ 
\bottomrule
\end{tabular}}
\end{table*}

\subsubsection{Fine-Grained Object Placement}
Beyond the generation of building structures, our system also reconstructs a variety of fine-grained urban elements to enrich the realism of the city scene. These include road-side objects such as street lamps, traffic signs, utility poles, benches, trash bins, and vegetation elements like trees or bushes. Since these objects are typically repetitive and exhibit limited geometric variation, we employ a \emph{retrieval-based} strategy instead of a fully generative one. This allows the system to efficiently populate the environment while preserving visual and semantic consistency with real cities.

Each object category is associated with a curated open-source 3D asset library that provides high-quality meshes and materials. Their spatial distribution is determined by two complementary placement mechanisms.  
(i) In the \textbf{rule-based placement}, spatial anchors are extracted from the road network obtained via OSM data. The algorithm detects road geometries and lane types (\texttt{primary}, \texttt{secondary}, or \texttt{service}) and places objects along road boundaries at regular intervals, with category-specific spacing, orientation, and offsets to ensure coherent alignment with the traffic infrastructure.  
(ii) In the \textbf{VLM-assisted placement}, VLMs are leveraged to interpret street-view imagery and textual tags. By analyzing geotagged images, the model infers both the semantic category and the most probable spatial location of contextual elements—for instance, determining where traffic lights or signposts appear relative to intersections or pedestrian crossings. Through the combination of structured geographic data and multimodal visual reasoning, our framework reconstructs not only the main urban geometry but also the rich layer of fine-grained objects that define the functional and aesthetic characteristics of real streetscapes.

\subsubsection{Road Network and Traffic Modeling}
After obtaining the initial coarse road network data from OSM, we integrate MOSS, a transportation simulator~\cite{zhang2024moss}, into our framework. This integration enables deriving a much finer and semantically richer representation of the urban road infrastructure and generating city-scale dynamic traffic, including both human and vehicle activities. The required assets for people and vehicles are sourced either by retrieving them from standard models or by generating them using Huanyuan-3D models. This refined process yields detailed lane-level topology and accurate connectivity between road segments, ensuring a faithful reconstruction of urban traffic structures. Leveraging these high-quality assets, our procedural modeling system then creates visually realistic and geometrically consistent road layouts that seamlessly integrate with surrounding urban environments. Crucially, the resulting 3D environment is simultaneously populated with large-scale dynamic traffic, providing crucial support for downstream applications, including embodied intelligence and world modeling.

\section{Experiments} \label{sec:exp}

Our evaluation is two-fold. First, we evaluate the quality and fidelity of 3D world generation using quantitative metrics and qualitative demonstrations. Second, we conduct comparison studies to validate our agentic design. To ensure fairness and comparability, identical regions, settings, and evaluation protocols are applied across all methods. Furthermore, experiments involving incommensurable paradigms are conducted under consistent sampling distributions to maintain equitable comparisons, as detailed below. Please also refer to the Supplementary Material for detailed implementation configurations and cost estimations.

\begin{figure*}[htbp]
    \centering
    \includegraphics[width=\textwidth]{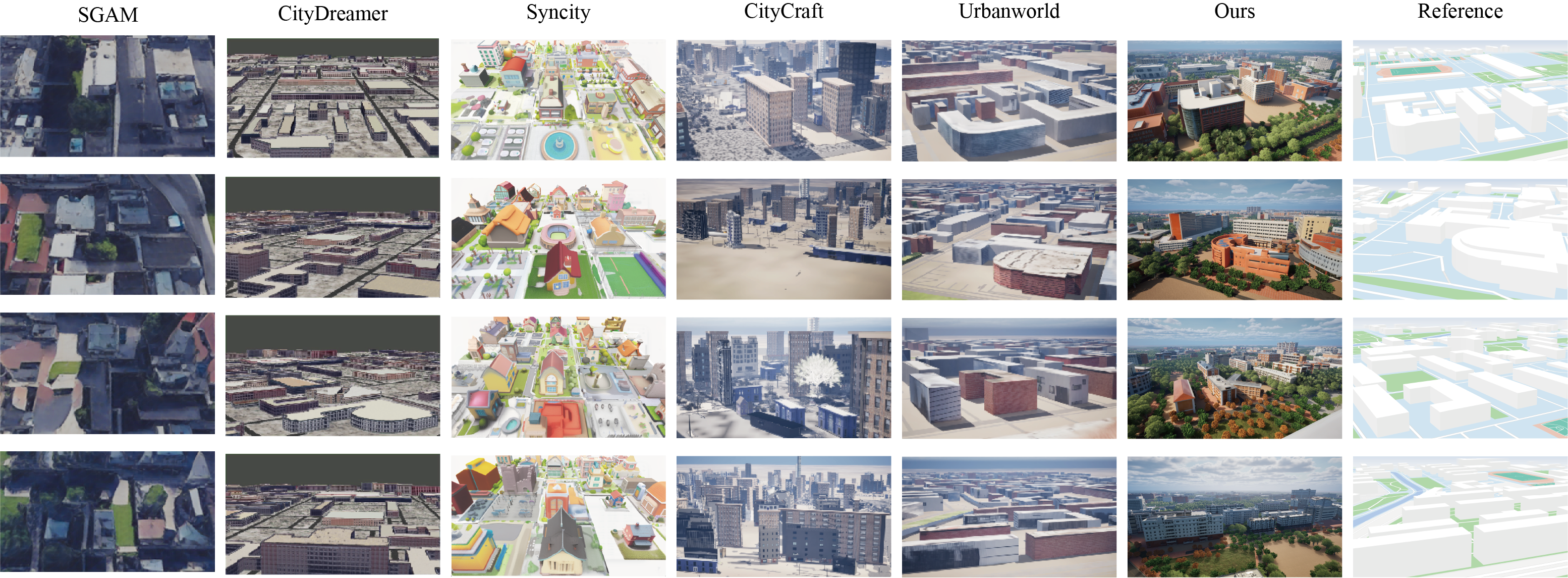}
    \caption{Qualitative comparison of methods; the final column shows a real-world scene from a commercial map.}
    \label{fig:compare_main}
\end{figure*}

\subsection{Quantitative 3D  World Evaluation}

Evaluating 3D scene generation is multifaceted, requiring assessment from large-scale layout accuracy to fine-grained visual quality. We conduct a comprehensive quantitative evaluation from two complementary perspectives: region-level layout and street-level quality.

\textbf{Setting and Metrics.} To assess large-scale layout accuracy, we employ the Learned Perceptual Image Patch Similarity (LPIPS)~\cite{zhang2018unreasonable} and Edge-IoU (E-IoU) to quantify the fidelity of the generated world layout against ground truth. Beyond layout, we evaluate the fine-grained, street-level quality of the generated 3D urban scenes. The LAION-Aesthetics Predictor (LAP)~\cite{schuhmann2022laion} is used to assess the aesthetic quality of the generated scenes. To capture human perceptual preferences, we follow common practice of llm-as-a-judge~\cite{gu2024survey,zheng2023judging} and employ GPT-5.4~\cite{openai2025gpt5}, a leading large vision-language model, as an evaluator for point-wise evaluation. Additionally, we conduct expert human evaluations for pairwise comparisons, which serves as the gold standard for assessing perceptual quality.

As shown in Table \ref{tab:quantitative_exp}, our method achieves highly competitive performance in layout alignment. Notably, while most baselines (e.g., all except CityCraft~\cite{deng2024citycraft}) are strictly constrained to OSM geometry, our more flexible approach meets or exceeds their performance, demonstrating high fidelity without rigid geometric priors.

The street-level evaluation results are also presented in Table~\ref{tab:quantitative_exp}. In a point-wise assessment evaluating geometric reasonability, texture quality, inter-object relations, overall visual effect, and fidelity, our method significantly surpasses all competitors. Furthermore, in direct pair-wise comparisons, scenes generated by \frameworkname achieved a win rate of over 86\% against all other methods, highlighting a distinct improvement in 3D urban world generation quality.

Further implementation details and additional analyzes are provided in the Supplementary Material.

\begin{table*}[htbp]
\centering
\caption{Quantitative comparison of 2D image quality and 3D reconstruction quality from different methods. For the 3D metrics, both Shape Coherence and Texture Quality are computed using expert curated streetview images as the reference. For FID/KID, lower is better ($\downarrow$), while for SSIM, CLIP-Sim., and Uni3D-I, higher is better ($\uparrow$). 
}
\label{tab:agent_ablation}
\resizebox{0.9\textwidth}{!}{
\begin{tabular}{l c cccc cc}
\toprule
\textbf{Method} & \textbf{Category} & \multicolumn{4}{c}{\textbf{2D Image Quality}} & \multicolumn{2}{c}{\textbf{3D Reconstruction}} \\
\cmidrule(lr){3-6} \cmidrule(lr){7-8}
& & & & & & \small{Shape Coherence} & \small{Texture Quality} \\
\cmidrule(lr){7-7} \cmidrule(lr){8-8}
& & FID $\downarrow$ & KID $\downarrow$ & SSIM $\uparrow$ & CLIP-Sim. $\uparrow$ & Uni3D-I $\uparrow$ & FID $\downarrow$ \\
\midrule
UrbanWorld & - & - & - & - & - & 0.1060 & 367.23 \\
\midrule

Text-Only & \multirow{2}{*}{Baselines} & 294.27 & 0.1612 & 0.3005 & 0.6527 & 0.0874 & 337.89 \\
Random Streetview & & 292.25 & 0.1610 & 0.2902 & 0.6466 & 0.0901 & 338.32 \\
\midrule

Expert Streetview & \multirow{2}{*}{Human} & \underline{292.09} & \underline{0.1529} & 0.2977 & \textbf{0.7272 }& \textbf{0.1067} & 338.21 \\
Expert Multi-Source & & 303.58 & 0.1696 & \textbf{0.3212} & 0.6947 & \underline{0.0991} & 341.70 \\
\midrule

Agent Streetview & \multirow{2}{*}{Agent} & \textbf{286.64} & \textbf{0.1413} & 0.2745 & \underline{0.6995} & 0.0951 & \textbf{315.74} \\
Agent Multi-Source & & 298.10 & 0.1538 & \underline{0.3033} & 0.6855 & 0.0937 & \underline{323.37} \\

\bottomrule
\end{tabular}}
\end{table*}

\subsection{Qualitative 3D World Evaluation}
 For a more holistic understanding, representative qualitative results are presented in Figure \ref{fig:compare_main}, visually illustrating the performance of our method against several baselines. Additional demonstrations of the generated 3D urban worlds, including videos, can be found in the Supplementary Material.

\textbf{Visualization Setting.} For mesh-based models, we render 3D assets with Unreal Engine 5~\cite{unrealengine} under identical lighting conditions, camera poses, and rendering parameters. For NeRF-based models, we follow the original implementations and render the scenes using closely matched camera poses to ensure visual comparability. The references are sampled from Amaps~\footnote{https://m.amap.com}, a leading provider of digital map in China.

The first column shows SGAM~\cite{shen2022sgam}. As an early large-scale attempt, it suffers from poor shape fidelity, low-resolution textures, and unrealistic spatial relations. Furthermore, its neural-based methodology imposes rigid viewpoint constraints, limiting applicability.
The second column presents CityDreamer~\cite{xie2024citydreamer}. Despite using OSM data, it yields overly simplified geometries and coarse textures. The method also struggles to incorporate auxiliary elements like vegetation, resulting in limited visual fidelity and aesthetic quality.
The third column illustrates Syncity~\cite{engstler2025syncity}, which stitches text-generated grids. Under our comparable evaluation, the results exhibit inconsistent realism and visible boundary discontinuities, despite reasonable local quality. Moreover, this strategy is inherently difficult to scale and struggles to incorporate fine-grained or dynamic elements.

The subsequent columns compare mesh-centric approaches. Although CityCraft~\cite{deng2024citycraft} generates high-precision models, its retrieval-based approach neglects road networks and spatial relationships, leading to unrealistic and conflicting layouts. Urbanworld~\cite{shang2024urbanworld} offers improved layout accuracy but is limited by coarse, primitive cuboid geometries and low-quality textures.

In contrast, our results, demonstrated in the penultimate column, show clear advantages. The novel design of \frameworkname yields significant improvements in building model precision, texture fidelity, and overall layout reasonableness and accuracy.

We utilize 3D scene data from a digital map as a high-fidelity reference for spatial layout accuracy. This data, curated for commercial services, is representative of a practical 3D urban world. However, its utility is specific: while object existence and location are accurate, the 3D models consist of coarse, untextured geometries. We employ this dataset as a benchmark to clearly demonstrate the spatial misalignment in competing baselines and to validate the superior performance of our approach.

\begin{figure}[htbp]
    \includegraphics[width=\linewidth]{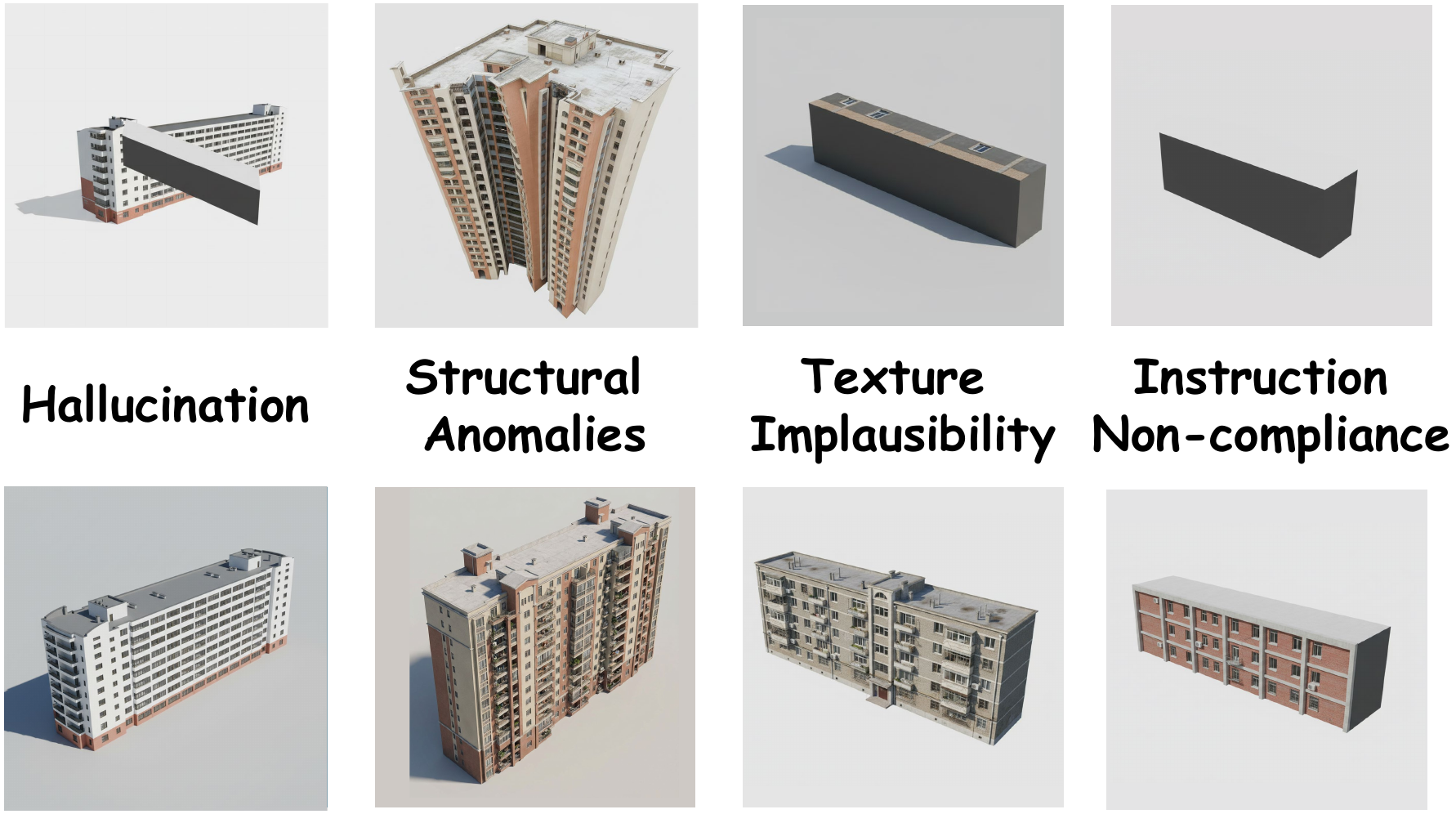}
    \caption{Four failure patterns observed during direct generation and how the \textit{Reflection} stage improves output quality. The first row displays the initially generated images, while the bottom row demonstrates the refined results following self-reflection and self-refinement.}
    \label{fig:failure}
    \vspace{-5pt}
\end{figure}

\begin{figure*}[htbp]
    \centering
    \includegraphics[width=\textwidth]{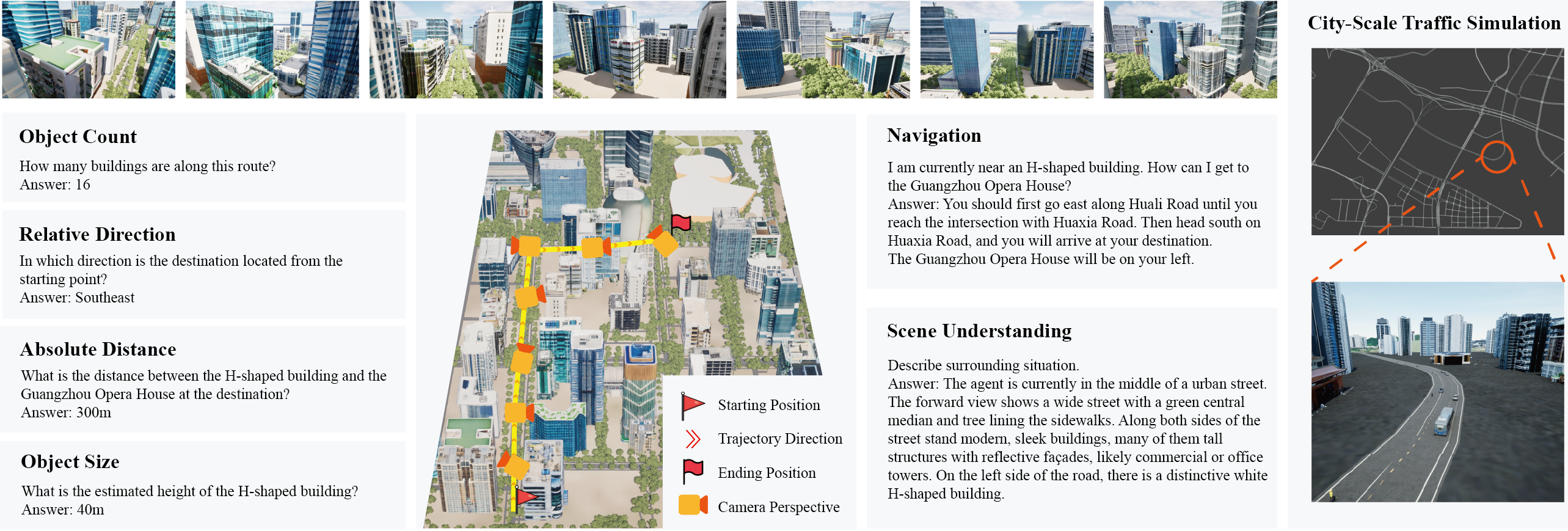}
    \caption{Downstream applications enabled by our framework.}
    \label{fig:app}
    \vspace{-5pt}
\end{figure*}

\subsection{\textit{Reflection} Mechanism Evaluation}

\textbf{Failure Analysis.} Although the \textit{Imagination} stage relies on capable foundation models, processing complex inputs can still introduce errors that compromise the quality of the constructed 3D scenes. We identify four primary failure patterns: \textbf{Hallucination}, defined as the generation of fabricated or unfaithful content; \textbf{Structural Anomalies}, where the generated buildings lack geometric or architectural plausibility; \textbf{Texture Implausibility}, in which the rendered surfaces appear artificial or implausible; and \textbf{Instruction Non-compliance}, where the model fails to adhere to the target generation instructions. Statistically, approximately $6\%$ of the samples score below $3$ (indicating ``Fair'' quality) in at least one dimension, which triggers the re-generation process. Further implementation details can be found in the Supplementary Material.

To validate the efficacy of this self-reflection and self-refinement design, we conduct a qualitative comparison before and after the \textit{Reflection} stage. As demonstrated in the bottom row of Figure~\ref{fig:failure}, the reflection loop effectively detects and mitigates the aforementioned errors. Specifically, it eliminates hallucinated artifacts to restore clean building envelopes, corrects structural anomalies into architecturally sound layouts, and replaces implausible textures with high-fidelity, realistic facades (e.g., adding detailed windows, balconies, and realistic lighting responses). Ultimately, this mechanism acts as a robust safeguard, ensuring the quality of the final 3D assets and minimizing cascade error accumulation.

\subsection{Autonomous Agent's Decision Evaluation}

\frameworkname utilizes an agentic design to effectively and efficiently select and process raw information from geographic information systems. In this section, we demonstrate that our agent's design is more effective for constructing urban 3D worlds than alternative methods, achieving performance that meets or exceeds that of human experts.

\textbf{Setting.} For this evaluation, we isolate the agent's decision-making by using fixed 2D imagination and 3D generation modules. We focus on the influence of input data selection and processing on the task of generating a single building's image and 3D object. We randomly sampled 50 buildings from the target region for our test set. A panel of human experts was invited to curate the corresponding standard street view images for these buildings, which serve as the ground-truth input data.

\textbf{Baselines and Metrics.} We evaluate two agent-based methods: (1) using street views selected and processed by our agent, and (2) using multi-source information (curated street view, OSM shape, volume) processed by our agent. We compare these approaches against four distinct baselines: (i) using only text descriptions, (ii) using randomly selected standard street views, (iii) using the human expert-curated ``golden" street views, and (iv) using standard multi-source information (expert curated street view, OSM shape, volume) without an agent. For the 2D image evaluation, we employ the Fréchet Inception Distance (FID)~\cite{heusel2017gans}, Kernel Inception Distance (KID)~\cite{binkowski2018demystifying}, SSIM~\cite{wang2004image}, and CLIP Similarity~\cite{radford2021learning}. For the 3D evaluation, we use Uni3D~\cite{zhou2023uni3d} to measure shape coherence and FID to assess texture quality.

The experimental results are presented in Table \ref{tab:agent_ablation}. The method utilizing street view images selected and processed autonomously by the agent demonstrates superior performance in perceptual quality. For 2D generation, it achieves the best image quality (as measured by FID and KID) and the second-highest semantic similarity (CLIP sim) among all methods. It also outperforms all other approaches in the texture quality (FID) of the 3D object construction. The methods incorporating multi-source data achieve the two highest SSIM scores, suggesting that explicit geometric information excels at structural alignment. The multi-source methods underperform their streetview-only counterparts on perceptual metrics. This suggests a trade-off in which rigid structural constraints may negatively impact visual fidelity. Overall, the agent-based methods outperform comparable non-agentic baselines across both 2D and 3D generation quality. 
Conversely, naive information retrieval methods (e.g., text-only or random views) perform poorly across all metrics. This result underscores the significant gap between raw data and high-quality, curated inputs, validating the importance of our intelligent agent design.

\subsection{Downstream Application}

With easily transform the real-world geospatial data and street-view images into 3D urban environment and related postprocessing suites, we can build diverse outdoors spatial reasoning tasks, navigation tasks and city-scale traffic simulation for any region. the potential downstream application enabled by our framework are presented in Figure~\ref{fig:app}. More details are presented in the corresponding section of the Supplementary Material.

\section{Related Work} \label{sec:related}

\textbf{3D Scene Generation.} Compared to generating 3D objects or avatars, generating 3D scenes presents significantly greater challenges~\cite{wen20253dscenegenerationsurvey}. The aim of 3D scene generation is to create a spatially structured, semantically meaningful, and visually realistic 3D environment for applications such as immersive media~\cite{anantrasirichai2022artificial,lee2024all,lavalle2023virtual,soliman2024artificial}, embodied intelligence~\cite{wang2024grutopia,wu2024metaurban,parish2001procedural,black2024pi_0}, and world models~\cite{ding2024understanding,liu2025generative,agarwal2025cosmos,zhu2024sora}. Procedural Generation~\cite{zhou2024scenex,yu2011make,musgrave1989synthesis,parish2001procedural,gu2025artiscene}, Neural 3D-based Generation~\cite{wang2018deep,xie2024citydreamer,lin2023infinicity}, Image-based Generation~\cite{dastjerdi2022guided,zhang2024text2nerf,yu2024wonderjourney}, and Video-based Generation~\cite{yu20244real,che2024gamegen,gao2024vista,gao2023magicdrive} are four major paradigms~\cite{wen20253dscenegenerationsurvey}. 
Existing 3D generation methodologies are characterized by significant trade-offs. Rule-based procedural generation~\cite{parish2001procedural}, while offering scalability and control, suffers from inflexibility and necessitates extensive human intervention. Concurrently, neural 3D methods are constrained by limited training data and suboptimal scalability, whereas visual-based approaches frequently exhibit deficiencies in geometric fidelity, view consistency, and compatibility with standard CG pipelines. By contrast, our multimodal framework facilitates scalable 3D urban generation characterized by enhanced reality alignment, photorealism, and view consistency. \frameworkname is distinctly based on real-world geospatial data, which differentiates it from purely imaginary 3D world generation frameworks~\cite{engstler2025syncity, hunyuanworldteam2025hunyuanworld10generatingimmersive}.

\textbf{Agent-based 3D generation.} Agent-based approaches in 3D synthesis leverage LLMs to plan, call tools, and verify outcomes across two granularities: \emph{object-level} (single asset creation and editing, ~\cite{chenIdea23DCollaborativeLMM2024, sun3DGPTProcedural3D2024, zhangShapeCraftLLMAgents2025}) and \emph{scene-level} (multi-object spatial layout and world construction, ~\cite{fengLayoutGPTCompositionalVisual2023, xia2025scenegenagent, yangSceneWeaverAllinOne3D2025, zhou2024scenex, zhangCityXControllableProcedural2024, tangUnrealLLMHighlyControllable2025}).
Idea23D~\cite{chenIdea23DCollaborativeLMM2024} coordinates multiple language–vision agents to translate mixed inputs (e.g., text, images, and optional 3D cues) into stepwise modeling operations, enabling iterative refinement of individual assets.
ShapeCraft~\cite{zhangShapeCraftLLMAgents2025} represents objects with a structured, program-like graph and employs LLM agents for parsing and incremental editing, supporting textured, editable, and interactive outputs. 
SceneWeaver~\cite{yangSceneWeaverAllinOne3D2025} adopts an extensible, self-reflective agent that selects appropriate scene-generation tools and performs automatic checks for semantic alignment, physical plausibility, and visual realism. 
UnrealLLM~\cite{tangUnrealLLMHighlyControllable2025} integrates LLM planning with Unreal Engine’s procedural content ecosystem, enabling high-level language control over asset retrieval, placement, and interactive editing. 
Collectively, these systems illustrate a progression from object-centric modeling to end-to-end scene and world construction with planning, tool use, and self-checking; unifying object- and scene-level reasoning within a single agentic framework remains an important open direction.

\section{Conclusion} \label{sec:conclusion}
In this paper, we propose \frameworkname, an agentic framework for generating reality-aligned 3D worlds at city-scale.
In this framework, an intelligent agent manages the 3D world generation by planning data and control flow, leveraging multimodal tools, and employing self-reflection for iterative refinement. 
\frameworkname outperforms existing 3D urban scene generation methods, featuring reality alignment, impressive 3D scene quality, view consistency, scalability, and seamless compatibility with existing CG pipelines. This highlights \frameworkname's significant potential for applications in embodied intelligence and world models.

\balance
\bibliographystyle{ACM-Reference-Format}
\bibliography{sample-base}


\begin{thebibliography}{65}


\ifx \showCODEN    \undefined \def \showCODEN     #1{\unskip}     \fi
\ifx \showISBNx    \undefined \def \showISBNx     #1{\unskip}     \fi
\ifx \showISBNxiii \undefined \def \showISBNxiii  #1{\unskip}     \fi
\ifx \showISSN     \undefined \def \showISSN      #1{\unskip}     \fi
\ifx \showLCCN     \undefined \def \showLCCN      #1{\unskip}     \fi
\ifx \shownote     \undefined \def \shownote      #1{#1}          \fi
\ifx \showarticletitle \undefined \def \showarticletitle #1{#1}   \fi
\ifx \showURL      \undefined \def \showURL       {\relax}        \fi
\providecommand\bibfield[2]{#2}
\providecommand\bibinfo[2]{#2}
\providecommand\natexlab[1]{#1}
\providecommand\showeprint[2][]{arXiv:#2}

\bibitem[Agarwal et~al\mbox{.}(2025)]%
        {agarwal2025cosmos}
\bibfield{author}{\bibinfo{person}{Niket Agarwal}, \bibinfo{person}{Arslan Ali}, \bibinfo{person}{Maciej Bala}, \bibinfo{person}{Yogesh Balaji}, \bibinfo{person}{Erik Barker}, \bibinfo{person}{Tiffany Cai}, \bibinfo{person}{Prithvijit Chattopadhyay}, \bibinfo{person}{Yongxin Chen}, \bibinfo{person}{Yin Cui}, \bibinfo{person}{Yifan Ding}, {et~al\mbox{.}}} \bibinfo{year}{2025}\natexlab{}.
\newblock \showarticletitle{Cosmos world foundation model platform for physical ai}.
\newblock \bibinfo{journal}{\emph{arXiv preprint arXiv:2501.03575}} (\bibinfo{year}{2025}).
\newblock


\bibitem[Anantrasirichai and Bull(2022)]%
        {anantrasirichai2022artificial}
\bibfield{author}{\bibinfo{person}{Nantheera Anantrasirichai} {and} \bibinfo{person}{David Bull}.} \bibinfo{year}{2022}\natexlab{}.
\newblock \showarticletitle{Artificial intelligence in the creative industries: a review}.
\newblock \bibinfo{journal}{\emph{Artificial intelligence review}} \bibinfo{volume}{55}, \bibinfo{number}{1} (\bibinfo{year}{2022}), \bibinfo{pages}{589--656}.
\newblock


\bibitem[Bi{\'n}kowski et~al\mbox{.}(2018)]%
        {binkowski2018demystifying}
\bibfield{author}{\bibinfo{person}{Miko{\l}aj Bi{\'n}kowski}, \bibinfo{person}{Danica~J Sutherland}, \bibinfo{person}{Michael Arbel}, {and} \bibinfo{person}{Arthur Gretton}.} \bibinfo{year}{2018}\natexlab{}.
\newblock \showarticletitle{Demystifying mmd gans}.
\newblock \bibinfo{journal}{\emph{arXiv preprint arXiv:1801.01401}} (\bibinfo{year}{2018}).
\newblock


\bibitem[Black et~al\mbox{.}(2024)]%
        {black2024pi_0}
\bibfield{author}{\bibinfo{person}{Kevin Black}, \bibinfo{person}{Noah Brown}, \bibinfo{person}{Danny Driess}, \bibinfo{person}{Adnan Esmail}, \bibinfo{person}{Michael Equi}, \bibinfo{person}{Chelsea Finn}, \bibinfo{person}{Niccolo Fusai}, \bibinfo{person}{Lachy Groom}, \bibinfo{person}{Karol Hausman}, \bibinfo{person}{Brian Ichter}, {et~al\mbox{.}}} \bibinfo{year}{2024}\natexlab{}.
\newblock \showarticletitle{$\backslash\pi_0$: A Vision-Language-Action Flow Model for General Robot Control}.
\newblock \bibinfo{journal}{\emph{arXiv preprint arXiv:2410.24164}} (\bibinfo{year}{2024}).
\newblock


\bibitem[Che et~al\mbox{.}(2024)]%
        {che2024gamegen}
\bibfield{author}{\bibinfo{person}{Haoxuan Che}, \bibinfo{person}{Xuanhua He}, \bibinfo{person}{Quande Liu}, \bibinfo{person}{Cheng Jin}, {and} \bibinfo{person}{Hao Chen}.} \bibinfo{year}{2024}\natexlab{}.
\newblock \showarticletitle{Gamegen-x: Interactive open-world game video generation}.
\newblock \bibinfo{journal}{\emph{arXiv preprint arXiv:2411.00769}} (\bibinfo{year}{2024}).
\newblock


\bibitem[Chen et~al\mbox{.}(2024)]%
        {chenIdea23DCollaborativeLMM2024}
\bibfield{author}{\bibinfo{person}{Junhao Chen}, \bibinfo{person}{Xiang Li}, \bibinfo{person}{Xiaojun Ye}, \bibinfo{person}{Chao Li}, \bibinfo{person}{Zhaoxin Fan}, {and} \bibinfo{person}{Hao Zhao}.} \bibinfo{year}{2024}\natexlab{}.
\newblock \bibinfo{title}{{{Idea23D}}: {{Collaborative LMM Agents Enable 3D Model Generation}} from {{Interleaved Multimodal Inputs}}}.
\newblock
\showeprint[arxiv]{2404.04363}~[cs]
\href{https://doi.org/10.48550/arXiv.2404.04363}{doi:\nolinkurl{10.48550/arXiv.2404.04363}}


\bibitem[Dastjerdi et~al\mbox{.}(2022)]%
        {dastjerdi2022guided}
\bibfield{author}{\bibinfo{person}{Mohammad Reza~Karimi Dastjerdi}, \bibinfo{person}{Yannick Hold-Geoffroy}, \bibinfo{person}{Jonathan Eisenmann}, \bibinfo{person}{Siavash Khodadadeh}, {and} \bibinfo{person}{Jean-Fran{\c{c}}ois Lalonde}.} \bibinfo{year}{2022}\natexlab{}.
\newblock \showarticletitle{Guided co-modulated gan for 360 field of view extrapolation}. In \bibinfo{booktitle}{\emph{2022 International Conference on 3D Vision (3DV)}}. IEEE, \bibinfo{pages}{475--485}.
\newblock


\bibitem[Deng et~al\mbox{.}(2024)]%
        {deng2024citycraft}
\bibfield{author}{\bibinfo{person}{Jie Deng}, \bibinfo{person}{Wenhao Chai}, \bibinfo{person}{Junsheng Huang}, \bibinfo{person}{Zhonghan Zhao}, \bibinfo{person}{Qixuan Huang}, \bibinfo{person}{Mingyan Gao}, \bibinfo{person}{Jianshu Guo}, \bibinfo{person}{Shengyu Hao}, \bibinfo{person}{Wenhao Hu}, \bibinfo{person}{Jenq-Neng Hwang}, {et~al\mbox{.}}} \bibinfo{year}{2024}\natexlab{}.
\newblock \showarticletitle{Citycraft: A real crafter for 3d city generation}.
\newblock \bibinfo{journal}{\emph{arXiv preprint arXiv:2406.04983}} (\bibinfo{year}{2024}).
\newblock


\bibitem[Ding et~al\mbox{.}(2024)]%
        {ding2024understanding}
\bibfield{author}{\bibinfo{person}{Jingtao Ding}, \bibinfo{person}{Yunke Zhang}, \bibinfo{person}{Yu Shang}, \bibinfo{person}{Yuheng Zhang}, \bibinfo{person}{Zefang Zong}, \bibinfo{person}{Jie Feng}, \bibinfo{person}{Yuan Yuan}, \bibinfo{person}{Hongyuan Su}, \bibinfo{person}{Nian Li}, \bibinfo{person}{Nicholas Sukiennik}, {et~al\mbox{.}}} \bibinfo{year}{2024}\natexlab{}.
\newblock \showarticletitle{Understanding world or predicting future? a comprehensive survey of world models}.
\newblock \bibinfo{journal}{\emph{Comput. Surveys}} (\bibinfo{year}{2024}).
\newblock


\bibitem[Engstler et~al\mbox{.}(2025)]%
        {engstler2025syncity}
\bibfield{author}{\bibinfo{person}{Paul Engstler}, \bibinfo{person}{Aleksandar Shtedritski}, \bibinfo{person}{Iro Laina}, \bibinfo{person}{Christian Rupprecht}, {and} \bibinfo{person}{Andrea Vedaldi}.} \bibinfo{year}{2025}\natexlab{}.
\newblock \showarticletitle{Syncity: Training-free generation of 3d worlds}.
\newblock \bibinfo{journal}{\emph{arXiv preprint arXiv:2503.16420}} (\bibinfo{year}{2025}).
\newblock


\bibitem[{Epic Games}(2025)]%
        {unrealengine}
\bibfield{author}{\bibinfo{person}{{Epic Games}}.} \bibinfo{year}{2025}\natexlab{}.
\newblock \bibinfo{title}{Unreal Engine}.
\newblock \bibinfo{howpublished}{\url{https://www.unrealengine.com}}.
\newblock


\bibitem[Feng et~al\mbox{.}(2023)]%
        {fengLayoutGPTCompositionalVisual2023}
\bibfield{author}{\bibinfo{person}{Weixi Feng}, \bibinfo{person}{Wanrong Zhu}, \bibinfo{person}{Tsu-jui Fu}, \bibinfo{person}{Varun Jampani}, \bibinfo{person}{Arjun Akula}, \bibinfo{person}{Xuehai He}, \bibinfo{person}{Sugato Basu}, \bibinfo{person}{Xin~Eric Wang}, {and} \bibinfo{person}{William~Yang Wang}.} \bibinfo{year}{2023}\natexlab{}.
\newblock \bibinfo{title}{{{LayoutGPT}}: {{Compositional Visual Planning}} and {{Generation}} with {{Large Language Models}}}.
\newblock
\showeprint[arxiv]{2305.15393}~[cs]
\href{https://doi.org/10.48550/arXiv.2305.15393}{doi:\nolinkurl{10.48550/arXiv.2305.15393}}


\bibitem[Gao et~al\mbox{.}(2023)]%
        {gao2023magicdrive}
\bibfield{author}{\bibinfo{person}{Ruiyuan Gao}, \bibinfo{person}{Kai Chen}, \bibinfo{person}{Enze Xie}, \bibinfo{person}{Lanqing Hong}, \bibinfo{person}{Zhenguo Li}, \bibinfo{person}{Dit-Yan Yeung}, {and} \bibinfo{person}{Qiang Xu}.} \bibinfo{year}{2023}\natexlab{}.
\newblock \showarticletitle{Magicdrive: Street view generation with diverse 3d geometry control}.
\newblock \bibinfo{journal}{\emph{arXiv preprint arXiv:2310.02601}} (\bibinfo{year}{2023}).
\newblock


\bibitem[Gao et~al\mbox{.}(2024)]%
        {gao2024vista}
\bibfield{author}{\bibinfo{person}{Shenyuan Gao}, \bibinfo{person}{Jiazhi Yang}, \bibinfo{person}{Li Chen}, \bibinfo{person}{Kashyap Chitta}, \bibinfo{person}{Yihang Qiu}, \bibinfo{person}{Andreas Geiger}, \bibinfo{person}{Jun Zhang}, {and} \bibinfo{person}{Hongyang Li}.} \bibinfo{year}{2024}\natexlab{}.
\newblock \showarticletitle{Vista: A generalizable driving world model with high fidelity and versatile controllability}.
\newblock \bibinfo{journal}{\emph{Advances in Neural Information Processing Systems}}  \bibinfo{volume}{37} (\bibinfo{year}{2024}), \bibinfo{pages}{91560--91596}.
\newblock


\bibitem[Garg(2025)]%
        {garg2025designing}
\bibfield{author}{\bibinfo{person}{Venus Garg}.} \bibinfo{year}{2025}\natexlab{}.
\newblock \showarticletitle{Designing the Mind: How Agentic Frameworks Are Shaping the Future of AI Behavior}.
\newblock \bibinfo{journal}{\emph{Journal of Computer Science and Technology Studies}} \bibinfo{volume}{7}, \bibinfo{number}{5} (\bibinfo{year}{2025}), \bibinfo{pages}{182--193}.
\newblock


\bibitem[{Google}(2025)]%
        {Google2025GeminiOverview}
\bibfield{author}{\bibinfo{person}{{Google}}.} \bibinfo{year}{2025}\natexlab{}.
\newblock \bibinfo{title}{Nano Banana: {Gemini} Image Generation Overview}.
\newblock
\urldef\tempurl%
\url{https://gemini.google/overview/image-generation/}
\showURL{%
\tempurl}


\bibitem[Gu et~al\mbox{.}(2024)]%
        {gu2024survey}
\bibfield{author}{\bibinfo{person}{Jiawei Gu}, \bibinfo{person}{Xuhui Jiang}, \bibinfo{person}{Zhichao Shi}, \bibinfo{person}{Hexiang Tan}, \bibinfo{person}{Xuehao Zhai}, \bibinfo{person}{Chengjin Xu}, \bibinfo{person}{Wei Li}, \bibinfo{person}{Yinghan Shen}, \bibinfo{person}{Shengjie Ma}, \bibinfo{person}{Honghao Liu}, {et~al\mbox{.}}} \bibinfo{year}{2024}\natexlab{}.
\newblock \showarticletitle{A survey on llm-as-a-judge}.
\newblock \bibinfo{journal}{\emph{arXiv preprint arXiv:2411.15594}} (\bibinfo{year}{2024}).
\newblock


\bibitem[Gu et~al\mbox{.}(2025)]%
        {gu2025artiscene}
\bibfield{author}{\bibinfo{person}{Zeqi Gu}, \bibinfo{person}{Yin Cui}, \bibinfo{person}{Zhaoshuo Li}, \bibinfo{person}{Fangyin Wei}, \bibinfo{person}{Yunhao Ge}, \bibinfo{person}{Jinwei Gu}, \bibinfo{person}{Ming-Yu Liu}, \bibinfo{person}{Abe Davis}, {and} \bibinfo{person}{Yifan Ding}.} \bibinfo{year}{2025}\natexlab{}.
\newblock \showarticletitle{Artiscene: Language-driven artistic 3d scene generation through image intermediary}. In \bibinfo{booktitle}{\emph{Proceedings of the Computer Vision and Pattern Recognition Conference}}. \bibinfo{pages}{2891--2901}.
\newblock


\bibitem[Heusel et~al\mbox{.}(2017)]%
        {heusel2017gans}
\bibfield{author}{\bibinfo{person}{Martin Heusel}, \bibinfo{person}{Hubert Ramsauer}, \bibinfo{person}{Thomas Unterthiner}, \bibinfo{person}{Bernhard Nessler}, {and} \bibinfo{person}{Sepp Hochreiter}.} \bibinfo{year}{2017}\natexlab{}.
\newblock \showarticletitle{Gans trained by a two time-scale update rule converge to a local nash equilibrium}.
\newblock \bibinfo{journal}{\emph{Advances in neural information processing systems}}  \bibinfo{volume}{30} (\bibinfo{year}{2017}).
\newblock


\bibitem[H{\"o}llein et~al\mbox{.}(2023)]%
        {hollein2023text2room}
\bibfield{author}{\bibinfo{person}{Lukas H{\"o}llein}, \bibinfo{person}{Ang Cao}, \bibinfo{person}{Andrew Owens}, \bibinfo{person}{Justin Johnson}, {and} \bibinfo{person}{Matthias Nie{\ss}ner}.} \bibinfo{year}{2023}\natexlab{}.
\newblock \showarticletitle{Text2room: Extracting textured 3d meshes from 2d text-to-image models}. In \bibinfo{booktitle}{\emph{Proceedings of the IEEE/CVF International Conference on Computer Vision}}. \bibinfo{pages}{7909--7920}.
\newblock


\bibitem[Hunyuan3D et~al\mbox{.}(2025)]%
        {hunyuan3d2025hunyuan3d}
\bibfield{author}{\bibinfo{person}{Team Hunyuan3D}, \bibinfo{person}{Shuhui Yang}, \bibinfo{person}{Mingxin Yang}, \bibinfo{person}{Yifei Feng}, \bibinfo{person}{Xin Huang}, \bibinfo{person}{Sheng Zhang}, \bibinfo{person}{Zebin He}, \bibinfo{person}{Di Luo}, \bibinfo{person}{Haolin Liu}, \bibinfo{person}{Yunfei Zhao}, {et~al\mbox{.}}} \bibinfo{year}{2025}\natexlab{}.
\newblock \showarticletitle{Hunyuan3D 2.1: From Images to High-Fidelity 3D Assets with Production-Ready PBR Material}.
\newblock \bibinfo{journal}{\emph{arXiv preprint arXiv:2506.15442}} (\bibinfo{year}{2025}).
\newblock


\bibitem[Kunda(2018)]%
        {kunda2018visual}
\bibfield{author}{\bibinfo{person}{Maithilee Kunda}.} \bibinfo{year}{2018}\natexlab{}.
\newblock \showarticletitle{Visual mental imagery: A view from artificial intelligence}.
\newblock \bibinfo{journal}{\emph{Cortex}}  \bibinfo{volume}{105} (\bibinfo{year}{2018}), \bibinfo{pages}{155--172}.
\newblock


\bibitem[LaValle(2023)]%
        {lavalle2023virtual}
\bibfield{author}{\bibinfo{person}{Steven~M LaValle}.} \bibinfo{year}{2023}\natexlab{}.
\newblock \bibinfo{booktitle}{\emph{Virtual reality}}.
\newblock \bibinfo{publisher}{Cambridge university press}.
\newblock


\bibitem[Lee et~al\mbox{.}(2024)]%
        {lee2024all}
\bibfield{author}{\bibinfo{person}{Lik-Hang Lee}, \bibinfo{person}{Tristan Braud}, \bibinfo{person}{Peng~Yuan Zhou}, \bibinfo{person}{Lin Wang}, \bibinfo{person}{Dianlei Xu}, \bibinfo{person}{Zijun Lin}, \bibinfo{person}{Abhishek Kumar}, \bibinfo{person}{Carlos Bermejo}, \bibinfo{person}{Pan Hui}, {et~al\mbox{.}}} \bibinfo{year}{2024}\natexlab{}.
\newblock \showarticletitle{All one needs to know about metaverse: A complete survey on technological singularity, virtual ecosystem, and research agenda}.
\newblock \bibinfo{journal}{\emph{Foundations and trends{\textregistered} in human-computer interaction}} \bibinfo{volume}{18}, \bibinfo{number}{2--3} (\bibinfo{year}{2024}), \bibinfo{pages}{100--337}.
\newblock


\bibitem[Lin et~al\mbox{.}(2023)]%
        {lin2023infinicity}
\bibfield{author}{\bibinfo{person}{Chieh~Hubert Lin}, \bibinfo{person}{Hsin-Ying Lee}, \bibinfo{person}{Willi Menapace}, \bibinfo{person}{Menglei Chai}, \bibinfo{person}{Aliaksandr Siarohin}, \bibinfo{person}{Ming-Hsuan Yang}, {and} \bibinfo{person}{Sergey Tulyakov}.} \bibinfo{year}{2023}\natexlab{}.
\newblock \showarticletitle{Infinicity: Infinite-scale city synthesis}. In \bibinfo{booktitle}{\emph{Proceedings of the IEEE/CVF international conference on computer vision}}. \bibinfo{pages}{22808--22818}.
\newblock


\bibitem[Liu et~al\mbox{.}(2025)]%
        {liu2025generative}
\bibfield{author}{\bibinfo{person}{Daochang Liu}, \bibinfo{person}{Junyu Zhang}, \bibinfo{person}{Anh-Dung Dinh}, \bibinfo{person}{Eunbyung Park}, \bibinfo{person}{Shichao Zhang}, \bibinfo{person}{Ajmal Mian}, \bibinfo{person}{Mubarak Shah}, {and} \bibinfo{person}{Chang Xu}.} \bibinfo{year}{2025}\natexlab{}.
\newblock \showarticletitle{Generative physical ai in vision: A survey}.
\newblock \bibinfo{journal}{\emph{arXiv preprint arXiv:2501.10928}} (\bibinfo{year}{2025}).
\newblock


\bibitem[Minderer et~al\mbox{.}(2022)]%
        {minderer2022simple}
\bibfield{author}{\bibinfo{person}{Matthias Minderer}, \bibinfo{person}{Alexey Gritsenko}, \bibinfo{person}{Austin Stone}, \bibinfo{person}{Maxim Neumann}, \bibinfo{person}{Dirk Weissenborn}, \bibinfo{person}{Alexey Dosovitskiy}, \bibinfo{person}{Aravindh Mahendran}, \bibinfo{person}{Anurag Arnab}, \bibinfo{person}{Mostafa Dehghani}, \bibinfo{person}{Zhuoran Shen}, {et~al\mbox{.}}} \bibinfo{year}{2022}\natexlab{}.
\newblock \showarticletitle{Simple open-vocabulary object detection}. In \bibinfo{booktitle}{\emph{European conference on computer vision}}. Springer, \bibinfo{pages}{728--755}.
\newblock


\bibitem[Musgrave et~al\mbox{.}(1989)]%
        {musgrave1989synthesis}
\bibfield{author}{\bibinfo{person}{F~Kenton Musgrave}, \bibinfo{person}{Craig~E Kolb}, {and} \bibinfo{person}{Robert~S Mace}.} \bibinfo{year}{1989}\natexlab{}.
\newblock \showarticletitle{The synthesis and rendering of eroded fractal terrains}.
\newblock \bibinfo{journal}{\emph{ACM Siggraph Computer Graphics}} \bibinfo{volume}{23}, \bibinfo{number}{3} (\bibinfo{year}{1989}), \bibinfo{pages}{41--50}.
\newblock


\bibitem[{OpenAI}(2025)]%
        {openai2025gpt5}
\bibfield{author}{\bibinfo{person}{{OpenAI}}.} \bibinfo{year}{2025}\natexlab{}.
\newblock \bibinfo{title}{Introducing GPT-5}.
\newblock \bibinfo{howpublished}{Web Page}.
\newblock
\urldef\tempurl%
\url{https://openai.com/blog/introducing-gpt-5}
\showURL{%
\tempurl}


\bibitem[{OpenStreetMap contributors}(2017)]%
        {OpenStreetMap}
\bibfield{author}{\bibinfo{person}{{OpenStreetMap contributors}}.} \bibinfo{year}{2017}\natexlab{}.
\newblock \bibinfo{title}{{Planet dump retrieved from https://planet.osm.org }}.
\newblock \bibinfo{howpublished}{\url{ https://www.openstreetmap.org }}.
\newblock


\bibitem[Parish and M{\"u}ller(2001)]%
        {parish2001procedural}
\bibfield{author}{\bibinfo{person}{Yoav~IH Parish} {and} \bibinfo{person}{Pascal M{\"u}ller}.} \bibinfo{year}{2001}\natexlab{}.
\newblock \showarticletitle{Procedural modeling of cities}. In \bibinfo{booktitle}{\emph{Proceedings of the 28th annual conference on Computer graphics and interactive techniques}}. \bibinfo{pages}{301--308}.
\newblock


\bibitem[Piao et~al\mbox{.}(2025)]%
        {piao2025agentsociety}
\bibfield{author}{\bibinfo{person}{Jinghua Piao}, \bibinfo{person}{Yuwei Yan}, \bibinfo{person}{Jun Zhang}, \bibinfo{person}{Nian Li}, \bibinfo{person}{Junbo Yan}, \bibinfo{person}{Xiaochong Lan}, \bibinfo{person}{Zhihong Lu}, \bibinfo{person}{Zhiheng Zheng}, \bibinfo{person}{Jing~Yi Wang}, \bibinfo{person}{Di Zhou}, {et~al\mbox{.}}} \bibinfo{year}{2025}\natexlab{}.
\newblock \showarticletitle{AgentSociety: Large-Scale Simulation of LLM-Driven Generative Agents Advances Understanding of Human Behaviors and Society}.
\newblock \bibinfo{journal}{\emph{arXiv preprint arXiv:2502.08691}} (\bibinfo{year}{2025}).
\newblock


\bibitem[Radford et~al\mbox{.}(2021)]%
        {radford2021learning}
\bibfield{author}{\bibinfo{person}{Alec Radford}, \bibinfo{person}{Jong~Wook Kim}, \bibinfo{person}{Chris Hallacy}, \bibinfo{person}{Aditya Ramesh}, \bibinfo{person}{Gabriel Goh}, \bibinfo{person}{Sandhini Agarwal}, \bibinfo{person}{Girish Sastry}, \bibinfo{person}{Amanda Askell}, \bibinfo{person}{Pamela Mishkin}, \bibinfo{person}{Jack Clark}, {et~al\mbox{.}}} \bibinfo{year}{2021}\natexlab{}.
\newblock \showarticletitle{Learning transferable visual models from natural language supervision}. In \bibinfo{booktitle}{\emph{International conference on machine learning}}. PmLR, \bibinfo{pages}{8748--8763}.
\newblock


\bibitem[Raistrick et~al\mbox{.}(2024)]%
        {raistrick2024infinigen}
\bibfield{author}{\bibinfo{person}{Alexander Raistrick}, \bibinfo{person}{Lingjie Mei}, \bibinfo{person}{Karhan Kayan}, \bibinfo{person}{David Yan}, \bibinfo{person}{Yiming Zuo}, \bibinfo{person}{Beining Han}, \bibinfo{person}{Hongyu Wen}, \bibinfo{person}{Meenal Parakh}, \bibinfo{person}{Stamatis Alexandropoulos}, \bibinfo{person}{Lahav Lipson}, {et~al\mbox{.}}} \bibinfo{year}{2024}\natexlab{}.
\newblock \showarticletitle{Infinigen indoors: Photorealistic indoor scenes using procedural generation}. In \bibinfo{booktitle}{\emph{Proceedings of the IEEE/CVF Conference on Computer Vision and Pattern Recognition}}. \bibinfo{pages}{21783--21794}.
\newblock


\bibitem[Schuhmann et~al\mbox{.}(2022)]%
        {schuhmann2022laion}
\bibfield{author}{\bibinfo{person}{Christoph Schuhmann}, \bibinfo{person}{Romain Beaumont}, \bibinfo{person}{Richard Vencu}, \bibinfo{person}{Cade Gordon}, \bibinfo{person}{Ross Wightman}, \bibinfo{person}{Mehdi Cherti}, \bibinfo{person}{Theo Coombes}, \bibinfo{person}{Aarush Katta}, \bibinfo{person}{Clayton Mullis}, \bibinfo{person}{Mitchell Wortsman}, {et~al\mbox{.}}} \bibinfo{year}{2022}\natexlab{}.
\newblock \showarticletitle{Laion-5b: An open large-scale dataset for training next generation image-text models}.
\newblock \bibinfo{journal}{\emph{Advances in neural information processing systems}}  \bibinfo{volume}{35} (\bibinfo{year}{2022}), \bibinfo{pages}{25278--25294}.
\newblock


\bibitem[Shang et~al\mbox{.}(2024)]%
        {shang2024urbanworld}
\bibfield{author}{\bibinfo{person}{Yu Shang}, \bibinfo{person}{Yuming Lin}, \bibinfo{person}{Yu Zheng}, \bibinfo{person}{Hangyu Fan}, \bibinfo{person}{Jingtao Ding}, \bibinfo{person}{Jie Feng}, \bibinfo{person}{Jiansheng Chen}, \bibinfo{person}{Li Tian}, {and} \bibinfo{person}{Yong Li}.} \bibinfo{year}{2024}\natexlab{}.
\newblock \showarticletitle{Urbanworld: An urban world model for 3d city generation}.
\newblock \bibinfo{journal}{\emph{arXiv preprint arXiv:2407.11965}} (\bibinfo{year}{2024}).
\newblock


\bibitem[Shen et~al\mbox{.}(2022)]%
        {shen2022sgam}
\bibfield{author}{\bibinfo{person}{Yuan Shen}, \bibinfo{person}{Wei-Chiu Ma}, {and} \bibinfo{person}{Shenlong Wang}.} \bibinfo{year}{2022}\natexlab{}.
\newblock \showarticletitle{SGAM: Building a virtual 3d world through simultaneous generation and mapping}.
\newblock \bibinfo{journal}{\emph{Advances in Neural Information Processing Systems}}  \bibinfo{volume}{35} (\bibinfo{year}{2022}), \bibinfo{pages}{22090--22102}.
\newblock


\bibitem[Shepard(1978)]%
        {shepard1978mental}
\bibfield{author}{\bibinfo{person}{Roger~N Shepard}.} \bibinfo{year}{1978}\natexlab{}.
\newblock \showarticletitle{The mental image.}
\newblock \bibinfo{journal}{\emph{American psychologist}} \bibinfo{volume}{33}, \bibinfo{number}{2} (\bibinfo{year}{1978}), \bibinfo{pages}{125}.
\newblock


\bibitem[Soliman et~al\mbox{.}(2024)]%
        {soliman2024artificial}
\bibfield{author}{\bibinfo{person}{Mona~M Soliman}, \bibinfo{person}{Eman Ahmed}, \bibinfo{person}{Ashraf Darwish}, {and} \bibinfo{person}{Aboul~Ella Hassanien}.} \bibinfo{year}{2024}\natexlab{}.
\newblock \showarticletitle{Artificial intelligence powered Metaverse: analysis, challenges and future perspectives}.
\newblock \bibinfo{journal}{\emph{Artificial Intelligence Review}} \bibinfo{volume}{57}, \bibinfo{number}{2} (\bibinfo{year}{2024}), \bibinfo{pages}{36}.
\newblock


\bibitem[Sun et~al\mbox{.}(2024)]%
        {sun3DGPTProcedural3D2024}
\bibfield{author}{\bibinfo{person}{Chunyi Sun}, \bibinfo{person}{Junlin Han}, \bibinfo{person}{Weijian Deng}, \bibinfo{person}{Xinlong Wang}, \bibinfo{person}{Zishan Qin}, {and} \bibinfo{person}{Stephen Gould}.} \bibinfo{year}{2024}\natexlab{}.
\newblock \bibinfo{title}{{{3D-GPT}}: {{Procedural 3D Modeling}} with {{Large Language Models}}}.
\newblock
\showeprint[arxiv]{2310.12945}~[cs]
\href{https://doi.org/10.48550/arXiv.2310.12945}{doi:\nolinkurl{10.48550/arXiv.2310.12945}}


\bibitem[Tang et~al\mbox{.}(2024)]%
        {tang2024diffuscene}
\bibfield{author}{\bibinfo{person}{Jiapeng Tang}, \bibinfo{person}{Yinyu Nie}, \bibinfo{person}{Lev Markhasin}, \bibinfo{person}{Angela Dai}, \bibinfo{person}{Justus Thies}, {and} \bibinfo{person}{Matthias Nie{\ss}ner}.} \bibinfo{year}{2024}\natexlab{}.
\newblock \showarticletitle{Diffuscene: Denoising diffusion models for generative indoor scene synthesis}. In \bibinfo{booktitle}{\emph{Proceedings of the IEEE/CVF conference on computer vision and pattern recognition}}. \bibinfo{pages}{20507--20518}.
\newblock


\bibitem[Tang et~al\mbox{.}(2025)]%
        {tangUnrealLLMHighlyControllable2025}
\bibfield{author}{\bibinfo{person}{Song Tang}, \bibinfo{person}{Kaiyong Zhao}, \bibinfo{person}{Lei Wang}, \bibinfo{person}{Yuliang Li}, \bibinfo{person}{Xuebo Liu}, \bibinfo{person}{Junyi Zou}, \bibinfo{person}{Qiang Wang}, {and} \bibinfo{person}{Xiaowen Chu}.} \bibinfo{year}{2025}\natexlab{}.
\newblock \showarticletitle{{{UnrealLLM}}: {{Towards Highly Controllable}} and {{Interactable 3D Scene Generation}} by {{LLM-powered Procedural Content Generation}}}. In \bibinfo{booktitle}{\emph{Findings of the {{Association}} for {{Computational Linguistics}}: {{ACL}} 2025}}, \bibfield{editor}{\bibinfo{person}{Wanxiang Che}, \bibinfo{person}{Joyce Nabende}, \bibinfo{person}{Ekaterina Shutova}, {and} \bibinfo{person}{Mohammad~Taher Pilehvar}} (Eds.). \bibinfo{publisher}{Association for Computational Linguistics}, \bibinfo{address}{Vienna, Austria}, \bibinfo{pages}{19417--19435}.
\newblock
\showISBNx{979-8-89176-256-5}
\href{https://doi.org/10.18653/v1/2025.findings-acl.994}{doi:\nolinkurl{10.18653/v1/2025.findings-acl.994}}


\bibitem[Team et~al\mbox{.}(2025)]%
        {hunyuanworldteam2025hunyuanworld10generatingimmersive}
\bibfield{author}{\bibinfo{person}{HunyuanWorld Team}, \bibinfo{person}{Zhenwei Wang}, \bibinfo{person}{Yuhao Liu}, \bibinfo{person}{Junta Wu}, \bibinfo{person}{Zixiao Gu}, \bibinfo{person}{Haoyuan Wang}, \bibinfo{person}{Xuhui Zuo}, \bibinfo{person}{Tianyu Huang}, \bibinfo{person}{Wenhuan Li}, \bibinfo{person}{Sheng Zhang}, \bibinfo{person}{Yihang Lian}, \bibinfo{person}{Yulin Tsai}, \bibinfo{person}{Lifu Wang}, \bibinfo{person}{Sicong Liu}, \bibinfo{person}{Puhua Jiang}, \bibinfo{person}{Xianghui Yang}, \bibinfo{person}{Dongyuan Guo}, \bibinfo{person}{Yixuan Tang}, \bibinfo{person}{Xinyue Mao}, \bibinfo{person}{Jiaao Yu}, \bibinfo{person}{Junlin Yu}, \bibinfo{person}{Jihong Zhang}, \bibinfo{person}{Meng Chen}, \bibinfo{person}{Liang Dong}, \bibinfo{person}{Yiwen Jia}, \bibinfo{person}{Chao Zhang}, \bibinfo{person}{Yonghao Tan}, \bibinfo{person}{Hao Zhang}, \bibinfo{person}{Zheng Ye}, \bibinfo{person}{Peng He}, \bibinfo{person}{Runzhou Wu}, \bibinfo{person}{Minghui Chen}, \bibinfo{person}{Zhan Li},
  \bibinfo{person}{Wangchen Qin}, \bibinfo{person}{Lei Wang}, \bibinfo{person}{Yifu Sun}, \bibinfo{person}{Lin Niu}, \bibinfo{person}{Xiang Yuan}, \bibinfo{person}{Xiaofeng Yang}, \bibinfo{person}{Yingping He}, \bibinfo{person}{Jie Xiao}, \bibinfo{person}{Yangyu Tao}, \bibinfo{person}{Jianchen Zhu}, \bibinfo{person}{Jinbao Xue}, \bibinfo{person}{Kai Liu}, \bibinfo{person}{Chongqing Zhao}, \bibinfo{person}{Xinming Wu}, \bibinfo{person}{Tian Liu}, \bibinfo{person}{Peng Chen}, \bibinfo{person}{Di Wang}, \bibinfo{person}{Yuhong Liu}, \bibinfo{person}{Linus}, \bibinfo{person}{Jie Jiang}, \bibinfo{person}{Tengfei Wang}, {and} \bibinfo{person}{Chunchao Guo}.} \bibinfo{year}{2025}\natexlab{}.
\newblock \bibinfo{title}{HunyuanWorld 1.0: Generating Immersive, Explorable, and Interactive 3D Worlds from Words or Pixels}.
\newblock
\showeprint[arxiv]{2507.21809}~[cs.CV]
\urldef\tempurl%
\url{https://arxiv.org/abs/2507.21809}
\showURL{%
\tempurl}


\bibitem[Wang et~al\mbox{.}(2024)]%
        {wang2024grutopia}
\bibfield{author}{\bibinfo{person}{Hanqing Wang}, \bibinfo{person}{Jiahe Chen}, \bibinfo{person}{Wensi Huang}, \bibinfo{person}{Qingwei Ben}, \bibinfo{person}{Tai Wang}, \bibinfo{person}{Boyu Mi}, \bibinfo{person}{Tao Huang}, \bibinfo{person}{Siheng Zhao}, \bibinfo{person}{Yilun Chen}, \bibinfo{person}{Sizhe Yang}, {et~al\mbox{.}}} \bibinfo{year}{2024}\natexlab{}.
\newblock \showarticletitle{Grutopia: Dream general robots in a city at scale}.
\newblock \bibinfo{journal}{\emph{arXiv preprint arXiv:2407.10943}} (\bibinfo{year}{2024}).
\newblock


\bibitem[Wang et~al\mbox{.}(2018)]%
        {wang2018deep}
\bibfield{author}{\bibinfo{person}{Kai Wang}, \bibinfo{person}{Manolis Savva}, \bibinfo{person}{Angel~X Chang}, {and} \bibinfo{person}{Daniel Ritchie}.} \bibinfo{year}{2018}\natexlab{}.
\newblock \showarticletitle{Deep convolutional priors for indoor scene synthesis}.
\newblock \bibinfo{journal}{\emph{ACM Transactions on Graphics (TOG)}} \bibinfo{volume}{37}, \bibinfo{number}{4} (\bibinfo{year}{2018}), \bibinfo{pages}{1--14}.
\newblock


\bibitem[Wang et~al\mbox{.}(2004)]%
        {wang2004image}
\bibfield{author}{\bibinfo{person}{Zhou Wang}, \bibinfo{person}{Alan~C Bovik}, \bibinfo{person}{Hamid~R Sheikh}, {and} \bibinfo{person}{Eero~P Simoncelli}.} \bibinfo{year}{2004}\natexlab{}.
\newblock \showarticletitle{Image quality assessment: from error visibility to structural similarity}.
\newblock \bibinfo{journal}{\emph{IEEE transactions on image processing}} \bibinfo{volume}{13}, \bibinfo{number}{4} (\bibinfo{year}{2004}), \bibinfo{pages}{600--612}.
\newblock


\bibitem[Wen et~al\mbox{.}(2025)]%
        {wen20253dscenegenerationsurvey}
\bibfield{author}{\bibinfo{person}{Beichen Wen}, \bibinfo{person}{Haozhe Xie}, \bibinfo{person}{Zhaoxi Chen}, \bibinfo{person}{Fangzhou Hong}, {and} \bibinfo{person}{Ziwei Liu}.} \bibinfo{year}{2025}\natexlab{}.
\newblock \bibinfo{title}{3D Scene Generation: A Survey}.
\newblock
\showeprint[arxiv]{2505.05474}~[cs.CV]
\urldef\tempurl%
\url{https://arxiv.org/abs/2505.05474}
\showURL{%
\tempurl}


\bibitem[Wu et~al\mbox{.}(2024)]%
        {wu2024metaurban}
\bibfield{author}{\bibinfo{person}{Wayne Wu}, \bibinfo{person}{Honglin He}, \bibinfo{person}{Yiran Wang}, \bibinfo{person}{Chenda Duan}, \bibinfo{person}{Jack He}, \bibinfo{person}{Zhizheng Liu}, \bibinfo{person}{Quanyi Li}, {and} \bibinfo{person}{Bolei Zhou}.} \bibinfo{year}{2024}\natexlab{}.
\newblock \showarticletitle{Metaurban: A simulation platform for embodied ai in urban spaces}.
\newblock \bibinfo{journal}{\emph{arXiv e-prints}} (\bibinfo{year}{2024}), \bibinfo{pages}{arXiv--2407}.
\newblock


\bibitem[Xia et~al\mbox{.}(2025)]%
        {xia2025scenegenagent}
\bibfield{author}{\bibinfo{person}{Xiao Xia}, \bibinfo{person}{Dan Zhang}, \bibinfo{person}{Zibo Liao}, \bibinfo{person}{Zhenyu Hou}, \bibinfo{person}{Tianrui Sun}, \bibinfo{person}{Jing Li}, \bibinfo{person}{Ling Fu}, {and} \bibinfo{person}{Yuxiao Dong}.} \bibinfo{year}{2025}\natexlab{}.
\newblock \showarticletitle{Scenegenagent: Precise industrial scene generation with coding agent}. In \bibinfo{booktitle}{\emph{Proceedings of the 63rd Annual Meeting of the Association for Computational Linguistics (Volume 1: Long Papers)}}. \bibinfo{pages}{17847--17875}.
\newblock


\bibitem[Xie et~al\mbox{.}(2024)]%
        {xie2024citydreamer}
\bibfield{author}{\bibinfo{person}{Haozhe Xie}, \bibinfo{person}{Zhaoxi Chen}, \bibinfo{person}{Fangzhou Hong}, {and} \bibinfo{person}{Ziwei Liu}.} \bibinfo{year}{2024}\natexlab{}.
\newblock \showarticletitle{Citydreamer: Compositional generative model of unbounded 3d cities}. In \bibinfo{booktitle}{\emph{Proceedings of the IEEE/CVF conference on computer vision and pattern recognition}}. \bibinfo{pages}{9666--9675}.
\newblock


\bibitem[Xie et~al\mbox{.}(2025)]%
        {xie2025citydreamer4d}
\bibfield{author}{\bibinfo{person}{Haozhe Xie}, \bibinfo{person}{Zhaoxi Chen}, \bibinfo{person}{Fangzhou Hong}, {and} \bibinfo{person}{Ziwei Liu}.} \bibinfo{year}{2025}\natexlab{}.
\newblock \showarticletitle{Compositional Generative Model of Unbounded 4{D} Cities}.
\newblock \bibinfo{journal}{\emph{IEEE Transactions on Pattern Analysis and Machine Intelligence}} (\bibinfo{year}{2025}).
\newblock
\href{https://doi.org/10.1109/TPAMI.2025.3603078}{doi:\nolinkurl{10.1109/TPAMI.2025.3603078}}


\bibitem[Yang et~al\mbox{.}(2025)]%
        {yangSceneWeaverAllinOne3D2025}
\bibfield{author}{\bibinfo{person}{Yandan Yang}, \bibinfo{person}{Baoxiong Jia}, \bibinfo{person}{Shujie Zhang}, {and} \bibinfo{person}{Siyuan Huang}.} \bibinfo{year}{2025}\natexlab{}.
\newblock \bibinfo{title}{{{SceneWeaver}}: {{All-in-One 3D Scene Synthesis}} with an {{Extensible}} and {{Self-Reflective Agent}}}.
\newblock
\showeprint[arxiv]{2509.20414}~[cs]
\href{https://doi.org/10.48550/arXiv.2509.20414}{doi:\nolinkurl{10.48550/arXiv.2509.20414}}


\bibitem[Yu et~al\mbox{.}(2024b)]%
        {yu20244real}
\bibfield{author}{\bibinfo{person}{Heng Yu}, \bibinfo{person}{Chaoyang Wang}, \bibinfo{person}{Peiye Zhuang}, \bibinfo{person}{Willi Menapace}, \bibinfo{person}{Aliaksandr Siarohin}, \bibinfo{person}{Junli Cao}, \bibinfo{person}{L{\'a}szl{\'o} Jeni}, \bibinfo{person}{Sergey Tulyakov}, {and} \bibinfo{person}{Hsin-Ying Lee}.} \bibinfo{year}{2024}\natexlab{b}.
\newblock \showarticletitle{4real: Towards photorealistic 4d scene generation via video diffusion models}.
\newblock \bibinfo{journal}{\emph{Advances in Neural Information Processing Systems}}  \bibinfo{volume}{37} (\bibinfo{year}{2024}), \bibinfo{pages}{45256--45280}.
\newblock


\bibitem[Yu et~al\mbox{.}(2024a)]%
        {yu2024wonderjourney}
\bibfield{author}{\bibinfo{person}{Hong-Xing Yu}, \bibinfo{person}{Haoyi Duan}, \bibinfo{person}{Junhwa Hur}, \bibinfo{person}{Kyle Sargent}, \bibinfo{person}{Michael Rubinstein}, \bibinfo{person}{William~T Freeman}, \bibinfo{person}{Forrester Cole}, \bibinfo{person}{Deqing Sun}, \bibinfo{person}{Noah Snavely}, \bibinfo{person}{Jiajun Wu}, {et~al\mbox{.}}} \bibinfo{year}{2024}\natexlab{a}.
\newblock \showarticletitle{Wonderjourney: Going from anywhere to everywhere}. In \bibinfo{booktitle}{\emph{Proceedings of the IEEE/CVF Conference on Computer Vision and Pattern Recognition}}. \bibinfo{pages}{6658--6667}.
\newblock


\bibitem[Yu et~al\mbox{.}(2011)]%
        {yu2011make}
\bibfield{author}{\bibinfo{person}{Lap-Fai Yu}, \bibinfo{person}{Sai~Kit Yeung}, \bibinfo{person}{Chi-Keung Tang}, \bibinfo{person}{Demetri Terzopoulos}, \bibinfo{person}{Tony~F Chan}, {and} \bibinfo{person}{Stanley~J Osher}.} \bibinfo{year}{2011}\natexlab{}.
\newblock \showarticletitle{Make it home: automatic optimization of furniture arrangement.}
\newblock \bibinfo{journal}{\emph{ACM Trans. Graph.}} \bibinfo{volume}{30}, \bibinfo{number}{4} (\bibinfo{year}{2011}), \bibinfo{pages}{86}.
\newblock


\bibitem[Zhang et~al\mbox{.}(2024a)]%
        {zhang2024moss}
\bibfield{author}{\bibinfo{person}{Jun Zhang}, \bibinfo{person}{Wenxuan Ao}, \bibinfo{person}{Junbo Yan}, \bibinfo{person}{Can Rong}, \bibinfo{person}{Depeng Jin}, \bibinfo{person}{Wei Wu}, {and} \bibinfo{person}{Yong Li}.} \bibinfo{year}{2024}\natexlab{a}.
\newblock \showarticletitle{Moss: A large-scale open microscopic traffic simulation system}.
\newblock \bibinfo{journal}{\emph{arXiv preprint arXiv:2405.12520}} (\bibinfo{year}{2024}).
\newblock


\bibitem[Zhang et~al\mbox{.}(2024b)]%
        {zhang2024text2nerf}
\bibfield{author}{\bibinfo{person}{Jingbo Zhang}, \bibinfo{person}{Xiaoyu Li}, \bibinfo{person}{Ziyu Wan}, \bibinfo{person}{Can Wang}, {and} \bibinfo{person}{Jing Liao}.} \bibinfo{year}{2024}\natexlab{b}.
\newblock \showarticletitle{Text2nerf: Text-driven 3d scene generation with neural radiance fields}.
\newblock \bibinfo{journal}{\emph{IEEE Transactions on Visualization and Computer Graphics}} \bibinfo{volume}{30}, \bibinfo{number}{12} (\bibinfo{year}{2024}), \bibinfo{pages}{7749--7762}.
\newblock


\bibitem[Zhang et~al\mbox{.}(2018)]%
        {zhang2018unreasonable}
\bibfield{author}{\bibinfo{person}{Richard Zhang}, \bibinfo{person}{Phillip Isola}, \bibinfo{person}{Alexei~A Efros}, \bibinfo{person}{Eli Shechtman}, {and} \bibinfo{person}{Oliver Wang}.} \bibinfo{year}{2018}\natexlab{}.
\newblock \showarticletitle{The unreasonable effectiveness of deep features as a perceptual metric}. In \bibinfo{booktitle}{\emph{Proceedings of the IEEE conference on computer vision and pattern recognition}}. \bibinfo{pages}{586--595}.
\newblock


\bibitem[Zhang et~al\mbox{.}(2025)]%
        {zhangShapeCraftLLMAgents2025}
\bibfield{author}{\bibinfo{person}{Shuyuan Zhang}, \bibinfo{person}{Chenhan Jiang}, \bibinfo{person}{Zuoou Li}, {and} \bibinfo{person}{Jiankang Deng}.} \bibinfo{year}{2025}\natexlab{}.
\newblock \bibinfo{title}{{{ShapeCraft}}: {{LLM Agents}} for {{Structured}}, {{Textured}} and {{Interactive 3D Modeling}}}.
\newblock
\showeprint[arxiv]{2510.17603}~[cs]
\href{https://doi.org/10.48550/arXiv.2510.17603}{doi:\nolinkurl{10.48550/arXiv.2510.17603}}


\bibitem[Zhang et~al\mbox{.}(2024c)]%
        {zhangCityXControllableProcedural2024}
\bibfield{author}{\bibinfo{person}{Shougao Zhang}, \bibinfo{person}{Mengqi Zhou}, \bibinfo{person}{Yuxi Wang}, \bibinfo{person}{Chuanchen Luo}, \bibinfo{person}{Rongyu Wang}, \bibinfo{person}{Yiwei Li}, \bibinfo{person}{Zhaoxiang Zhang}, {and} \bibinfo{person}{Junran Peng}.} \bibinfo{year}{2024}\natexlab{c}.
\newblock \bibinfo{title}{{{CityX}}: {{Controllable Procedural Content Generation}} for {{Unbounded 3D Cities}}}.
\newblock
\showeprint[arxiv]{2407.17572}~[cs]
\href{https://doi.org/10.48550/arXiv.2407.17572}{doi:\nolinkurl{10.48550/arXiv.2407.17572}}


\bibitem[Zheng et~al\mbox{.}(2023)]%
        {zheng2023judging}
\bibfield{author}{\bibinfo{person}{Lianmin Zheng}, \bibinfo{person}{Wei-Lin Chiang}, \bibinfo{person}{Ying Sheng}, \bibinfo{person}{Siyuan Zhuang}, \bibinfo{person}{Zhanghao Wu}, \bibinfo{person}{Yonghao Zhuang}, \bibinfo{person}{Zi Lin}, \bibinfo{person}{Zhuohan Li}, \bibinfo{person}{Dacheng Li}, \bibinfo{person}{Eric Xing}, {et~al\mbox{.}}} \bibinfo{year}{2023}\natexlab{}.
\newblock \showarticletitle{Judging llm-as-a-judge with mt-bench and chatbot arena}.
\newblock \bibinfo{journal}{\emph{Advances in neural information processing systems}}  \bibinfo{volume}{36} (\bibinfo{year}{2023}), \bibinfo{pages}{46595--46623}.
\newblock


\bibitem[Zhou et~al\mbox{.}(2023)]%
        {zhou2023uni3d}
\bibfield{author}{\bibinfo{person}{Junsheng Zhou}, \bibinfo{person}{Jinsheng Wang}, \bibinfo{person}{Baorui Ma}, \bibinfo{person}{Yu-Shen Liu}, \bibinfo{person}{Tiejun Huang}, {and} \bibinfo{person}{Xinlong Wang}.} \bibinfo{year}{2023}\natexlab{}.
\newblock \showarticletitle{Uni3d: Exploring unified 3d representation at scale}.
\newblock \bibinfo{journal}{\emph{arXiv preprint arXiv:2310.06773}} (\bibinfo{year}{2023}).
\newblock


\bibitem[Zhou et~al\mbox{.}(2024)]%
        {zhou2024scenex}
\bibfield{author}{\bibinfo{person}{Mengqi Zhou}, \bibinfo{person}{Yuxi Wang}, \bibinfo{person}{Jun Hou}, \bibinfo{person}{Shougao Zhang}, \bibinfo{person}{Yiwei Li}, \bibinfo{person}{Chuanchen Luo}, \bibinfo{person}{Junran Peng}, {and} \bibinfo{person}{Zhaoxiang Zhang}.} \bibinfo{year}{2024}\natexlab{}.
\newblock \showarticletitle{SceneX: Procedural Controllable Large-scale Scene Generation}.
\newblock \bibinfo{journal}{\emph{arXiv preprint arXiv:2403.15698}} (\bibinfo{year}{2024}).
\newblock


\bibitem[Zhou et~al\mbox{.}(2025)]%
        {zhou2025virtualcommunityopenworld}
\bibfield{author}{\bibinfo{person}{Qinhong Zhou}, \bibinfo{person}{Hongxin Zhang}, \bibinfo{person}{Xiangye Lin}, \bibinfo{person}{Zheyuan Zhang}, \bibinfo{person}{Yutian Chen}, \bibinfo{person}{Wenjun Liu}, \bibinfo{person}{Zunzhe Zhang}, \bibinfo{person}{Sunli Chen}, \bibinfo{person}{Lixing Fang}, \bibinfo{person}{Qiushi Lyu}, \bibinfo{person}{Xinyu Sun}, \bibinfo{person}{Jincheng Yang}, \bibinfo{person}{Zeyuan Wang}, \bibinfo{person}{Bao~Chi Dang}, \bibinfo{person}{Zhehuan Chen}, \bibinfo{person}{Daksha Ladia}, \bibinfo{person}{Jiageng Liu}, {and} \bibinfo{person}{Chuang Gan}.} \bibinfo{year}{2025}\natexlab{}.
\newblock \bibinfo{title}{Virtual Community: An Open World for Humans, Robots, and Society}.
\newblock
\showeprint[arxiv]{2508.14893}~[cs.CV]
\urldef\tempurl%
\url{https://arxiv.org/abs/2508.14893}
\showURL{%
\tempurl}


\bibitem[Zhu et~al\mbox{.}(2024)]%
        {zhu2024sora}
\bibfield{author}{\bibinfo{person}{Zheng Zhu}, \bibinfo{person}{Xiaofeng Wang}, \bibinfo{person}{Wangbo Zhao}, \bibinfo{person}{Chen Min}, \bibinfo{person}{Nianchen Deng}, \bibinfo{person}{Min Dou}, \bibinfo{person}{Yuqi Wang}, \bibinfo{person}{Botian Shi}, \bibinfo{person}{Kai Wang}, \bibinfo{person}{Chi Zhang}, {et~al\mbox{.}}} \bibinfo{year}{2024}\natexlab{}.
\newblock \showarticletitle{Is sora a world simulator? a comprehensive survey on general world models and beyond}.
\newblock \bibinfo{journal}{\emph{arXiv preprint arXiv:2405.03520}} (\bibinfo{year}{2024}).
\newblock


\end{thebibliography}

\end{document}